\documentclass[final,3p,times]{elsarticle}
\usepackage{amssymb}
\usepackage{amsmath}
\usepackage{multirow}
\usepackage{booktabs}
\usepackage{balance}
\usepackage{colortbl}

\usepackage[colorlinks,
linkcolor=blue,
anchorcolor=blue,
citecolor=blue]{hyperref}

\hypersetup{hidelinks,
	colorlinks=true,
	allcolors=blue,
	pdfstartview=Fit,
	breaklinks=true}

\journal{}

\begin{document}

\begin{frontmatter}



\title{Joint Masked Reconstruction and Contrastive Learning for Mining Interactions Between Proteins}

\author{Jiang Li\fnref{label1,label2,label3,label4}}
\ead{lijfrank@hust.edu.cn}
\affiliation[label1]{organization={School of Artificial Intelligence and Automation, Huazhong University of Science and Technology},
            city={Wuhan},
            postcode={430074},
            country={China}}
\affiliation[label2]{organization={Institute of Artificial Intelligence, Huazhong University of Science and Technology}, 
            city={Wuhan},
            postcode={430074}, 
            country={China}}
\affiliation[label3]{organization={Hubei Key Laboratory of Brain-Inspired Intelligent Systems, Huazhong University of Science and Technology}, 
            city={Wuhan},
            postcode={430074}, 
            country={China}}
\affiliation[label4]{organization={Key Laboratory of Image Processing and Intelligent Control (Huazhong University of Science and Technology), Ministry of Education}, 
            city={Wuhan},
            postcode={430074}, 
            country={China}}

\author{Xiaoping Wang\corref{cor1}\fnref{label1,label3,label4}} %% Author name
\cortext[cor1]{Corresponding author.}
\ead{wangxiaoping@hust.edu.cn}
\fntext[]{Received 6 March 2025; Received in revised form 6 March 2025; Accepted 6 March 2025}
%% Abstract
\begin{abstract}
Protein-protein interaction (PPI) prediction is an instrumental means in elucidating the mechanisms underlying cellular operations, holding significant practical implications for the realms of pharmaceutical development and clinical treatment. Presently, the majority of research methods primarily concentrate on the analysis of amino acid sequences, while investigations predicated on protein structures remain in the nascent stages of exploration. Despite the emergence of several structure-based algorithms in recent years, these are still confronted with inherent challenges: (1) the extraction of intrinsic structural information of proteins typically necessitates the expenditure of substantial computational resources; (2) these models are overly reliant on seen protein data, struggling to effectively unearth interaction cues between unknown proteins. To further propel advancements in this domain, this paper introduces a novel PPI prediction method jointing masked reconstruction and contrastive learning, termed JmcPPI. This methodology dissects the PPI prediction task into two distinct phases: during the residue structure encoding phase, JmcPPI devises two feature reconstruction tasks and employs graph attention mechanism to capture structural information between residues; during the protein interaction inference phase, JmcPPI perturbs the original PPI graph and employs a multi-graph contrastive learning strategy to thoroughly mine extrinsic interaction information of novel proteins. Extensive experiments conducted on three widely utilized PPI datasets demonstrate that JmcPPI surpasses existing optimal baseline models across various data partition schemes. The associated code can be accessed via https://github.com/lijfrank-open/JmcPPI.
\end{abstract}

% \includegraphics[width=\linewidth]{overall_workflow.pdf}

% \begin{highlights}
% \item This study converts PPI prediction into two stages and proposes a new structure-based PPI prediction method.
% \item The proposed method utilizes the graph attention mechanism to accurately mine the structural features of proteins in the residue reconstruction stage.
% \item In the interaction prediction stage, the proposed method integrates two auxiliary tasks to enhance the protein expression capability of the encoder.
% \end{highlights}

%% Keywords

\end{frontmatter}

% \linenumbers

\section{Introduction}\label{sec:Introduction}
Proteins exhibit indispensable biological significance by participating in various physiological processes within organisms, such as metabolism and reproduction. Protein-protein interactions (PPIs) serve as the central hub of molecular mechanisms, directly regulating key biological events such as the formation of protein complexes, modulation of enzymatic activity, and signal transduction pathways. Accurate dissection of PPI networks not only provides a molecular blueprint for elucidating the pathogenesis of diseases but also exerts a profound impact on the development of targeted drugs, clinical diagnostics, and applications in synthetic biology. However, systematically deciphering functionally relevant patterns from the intricately interwoven networks still faces significant technological challenges. Moreover, the innovation of high-throughput sequencing technologies has driven an exponential expansion of the protein sequence space, yet the rate of functional annotation lags behind the pace of sequence discovery, leading to a persistent widening of the sequence-function annotation gap~\cite{10049745}. To surmount the limitations of traditional wet-lab approaches~\cite{Fields1994The, Zhu2001Global, Ho2002Systematic, Gavin2002Functional} in terms of throughput, cost, and scalability, the computational biology community has been continuously committed to developing machine learning-based PPI prediction frameworks.

Early research is predominantly founded upon the framework of statistical machine learning, with representative methods encompassing support vector machine~\cite{ijms17010021, 7454750}, random forest~\cite{You2015Predicting, 10.1093/bioinformatics/btx005}, conditional random field~\cite{10.1093/bioinformatics/btl660}, and naive Bayes~\cite{10.1093/bioinformatics/btq302, 10.1002/pmic.201200326}. These methodologies are predicated on the positive correlation between sequence homology and binding affinity, inferring interactions through manually designed sequence features. However, such shallow models are constrained by an insufficiency in nonlinear representational capacity and a dependency on manual feature engineering, rendering them incapable of capturing higher-order topological characteristics within PPI networks. The advent of deep learning technology signifies a paradigm shift in methodology, achieving end-to-end feature learning through multilayer nonlinear transformations and markedly enhancing the modeling capabilities of complex biological relationships. Li et al.~\cite{Li2022SDNN-PPI} designed self-attention deep neural networks, enhancing model interpretability through multi-scale feature extraction strategies such as amino acid composition, combined triplet, and auto-covariance. Hu et al.~\cite{10.1093/bioinformatics/btab737} introduced a single-protein category recognition module, combining multi-scale convolutional neural networks (CNNs) to parse contextual semantic information from protein sequences. These groundbreaking advancements~\cite{Li2022SDNN-PPI,10.1093/bioinformatics/btab737,Sun2017Sequence-based,10.7717/peerj.7126,dutta-saha-2020-amalgamation} collectively propelled the evolution of PPI prediction from shallow statistical models to deep representation learning paradigms.
In recent years, the research paradigm has gradually shifted towards modeling protein interactions based on graph neural networks (GNNs), significantly enhancing predictive performance by integrating the topological correlations between proteins. Zhao et al.~\cite{zhao2023semignn} further proposed a self-ensembling multi-graph network, achieving multi-type PPI prediction by simulating protein correlations and label dependencies. Tang et al.~\cite{10.1093/bioinformatics/btae603} employed an antisymmetric graph network architecture to model the heterogeneous distribution within PPI networks and optimized the weight allocation of different types of PPIs using asymmetric loss functions. Despite the significant advancements~\cite{zhao2023semignn,10.1093/bioinformatics/btae603,Huang2020SkipGNN,10.1093/bib/bbab513,ijcai2021p506} achieved by these methods in sequence-driven PPI prediction, the realization of protein function fundamentally relies on the folded conformation of its tertiary structure. Therefore, prediction methods that solely depend on sequence information still have inherent limitations when modeling complex PPI networks, failing to fully reflect the actual interaction mechanisms of proteins in three-dimensional space.

To transcend the limitations of sequence-driven approaches, a research paradigm based on protein structure has emerged, focusing on leveraging the relational modeling capabilities of GNNs to mine the internal structural information of proteins. Gao et al.~\cite{gao2023hierarchical} proposed a dual-view hierarchical graph network framework, with the lower-layer graph network mining protein structural information and the upper-layer graph network capturing PPI interaction features, achieving multi-level structure-aware modeling. Huang et al.~\cite{10.1093/bib/bbad020} not only considered the overall and local structural features of proteins but also incorporated evolutionary profiles into PPI structural representation to enhance model performance. However, these structure-based methods~\cite{gao2023hierarchical,10.1093/bib/bbad020,Baranwal2022Struct2Graph,Jha2022Prediction} are still in the early stages of exploration, and their core challenges are (1) high computational resource consumption: the need for in-depth mining of internal structural information for each protein leads to high computational costs in model training and inference processes; (2) limited generalization ability: existing methods are overly reliant on known protein structural information and struggle to effectively predict interactions between novel proteins. To address these issues, Wu et al.~\cite{wu2024mapeppi} proposed the MAPE-PPI framework, which breaks down PPI prediction into two stages, i.e., microenvironment codebook learning and protein encoding, attempting to reduce computational complexity. However, their method still has the following limitations:

\begin{enumerate}[(1)]
     \item Redundant computation issue -- During the microenvironment codebook learning phase, MAPE-PPI simultaneously trains the graph encoder and microenvironment codebook, leading to high memory resource consumption. In fact, the graph encoder can effectively extract residue microenvironment information through the message passing mechanism, making additional codebook training redundant.
     \item Insufficient modeling of neighboring residue influence -- MAPE-PPI uses standard graph convolutional network (GCN)~\cite{kipf2017semisupervised} for information transmission, treating the influence of different neighboring residues as equivalent. This way lacks adaptive modeling capabilities for local structural differences, potentially affecting the accurate extraction of internal protein structural information.
     \item Limited prediction ability for novel proteins -- Similar to existing graph-based methods~\cite{zhao2023semignn, ijcai2021p506, gao2023hierarchical}, MAPE-PPI uses conventional graph structure framework to model PPI networks, and its predictive performance heavily relies on interaction patterns of previously seen proteins. This may hinder the full exploration of potential interaction clues between novel proteins, leading to insufficient generalization capacity.
\end{enumerate}

To address the aforementioned challenges, in this work, we propose joint masked reconstruction and contrastive learning for PPI (JmcPPI) prediction method. The proposed method employs GNNs to mine both the internal structural cues and external interaction cues of proteins, encompassing two principal phases, i.e., residue structure encoding and protein interaction inference. During the residue structure encoding phase, we construct a heterogeneous protein graph that includes three types of relational connections and utilize HetGNN~\cite{10.1145/3292500.3330961} as the residue structure encoder to capture the topological structural information between residues. To enhance the robustness of the encoder, we design a masked reconstruction task that shares parameters with the standard reconstruction task. This task involves randomly masking some residue features, compelling the model to learn context-aware representations. Additionally, we incorporate the GAT~\cite{veličković2018graph} operator within the HetGNN for message passing to dynamically compute the aggregation weights of residues. At this stage, the process of GAT aggregating neighboring information effectively integrates the microenvironment, obviating the redundant computation of microenvironment codebooks as seen in MAPE-PPI. In the protein interaction inference phase, we construct a protein interaction graph based on the PPI network and employ GIN~\cite{xu2018how} to mine the interaction information between proteins. To further extract interaction cues between novel proteins, we apply a certain rate of perturbation to the original interaction graph to generate two new graphs and use encoders with shared parameters for message propagation. Building on this, we devise a multi-graph contrastive learning task to constrain the similarity of positive-negative sample pairs in the latent representation space. In this manner, we can compel the protein encoder to infer the potential representations of proteins from limited information, significantly reducing the model's overfitting tendency to features of previously seen proteins and thereby fully extracting interaction cues between novel proteins. We conduct extensive experiments on three public PPI datasets, and the results demonstrate that JmcPPI achieves optimal performance under most data partition schemes. To put it in a nutshell, the contributions of this paper can be encapsulated as follows:
\begin{enumerate}[(1)]
     \item We propose JmcPPI, the first PPI prediction method that integrates residue masking reconstruction and multi-graph contrastive learning, which is capable of capturing both structural and interactional cues of proteins while alleviating the issue of high resource consumption through task decomposition.
     \item During the residue structure encoding phase, we combine heterogeneous graph network with attention mechanism and design two reconstruction tasks to fully mine the internal structural information of proteins.
     \item In the protein interaction inference phase, we generate two new interaction graphs and devise a multi-graph contrastive learning task, compelling the encoder to thoroughly mine the external interaction information of novel proteins.
     \item We perform a multitude of experiments on various PPI datasets, including model comparison, generalizability analysis, and ablation study. The outcomes substantiate the superior performance of JmcPPI over state-of-the-art baselines across diverse data partitions.
\end{enumerate}

The organization of this paper is as follows: Section~\ref{sec:Related_Work} reviews the relevant research on PPI prediction, elucidating the advancements and limitations of existing methodologies; Section~\ref{sec:Methodology} provides a detailed exposition of the proposed JmcPPI method, encompassing its underlying principles, technical framework, and model details; Section~\ref{sec:Experimental_Setup} and Section~\ref{sec:Experimental_Result} respectively introduce the experimental setup and results analysis, substantiating the effectiveness and superiority of the proposed method through extensive experimentation; finally, Section~\ref{sec:Conclusion} summarizes the research endeavors of this paper and projects potential directions for future research.

\section{Related Work}\label{sec:Related_Work}
The dynamic regulation of cellular functions is contingent upon the intricate network of protein interactions, and such molecular interaction mechanisms hold significant value for elucidating disease pathogenesis and the development of targeted drugs~\cite{Keskin2008Characterization, Hakes2008Protein, Wang2022Protein}. Current research indicates a strong correlation between the PPI network's topological properties and protein functions, with the connection patterns between nodes providing crucial clues for the identification of disease biomarkers. Early studies~\cite{Guo2008Using, Zhou2017Multi-scale, 10.1002/prot.26486} on PPI prediction primarily relied on traditional machine learning models (such as support vector machines~\cite{Chatterjee2011PPI_SVM}, random forests~\cite{You2015Predicting}, etc.) for predictions, but these methods were constrained by model complexity and struggled to model the nonlinear associations between proteins, with the feature engineering process requiring manual intervention. With the advancement of deep learning technology, prediction frameworks based on neural networks~\cite{Sun2017Sequence-based, Li2022SDNN-PPI, dutta-saha-2020-amalgamation} have significantly enhanced performance through automatic feature learning. Typical works include hybrid architectures combining residual convolution and recurrent neural networks~\cite{10.1093/bioinformatics/btz328}, models integrating data augmentation strategies~\cite{10.1093/bioinformatics/bty573}, and methods introducing attention mechanism~\cite{asim2022adhppi}. Sun et al.~\cite{Sun2017Sequence-based} pioneered an unsupervised sequence feature extraction framework based on stacked autoencoders, forging a new path for the application of deep learning in PPI prediction. Yao et al.~\cite{10.7717/peerj.7126}, inspired by natural language processing, developed the Res2vec residue embedding algorithm, which embedded residues into low-dimensional vectors for protein sequence representation. Dutta et al.~\cite{dutta-saha-2020-amalgamation} were the first to construct a multimodal deep learning architecture, integrating textual descriptions, three-dimensional structures, and genomic sequences to achieve cross-modality feature fusion. Although these approaches have broken through the limitations of shallow models, the utilization of structural information such as protein three-dimensional conformations remains insufficient.

In recent years, PPI prediction methods based on graph representation learning~\cite{Huang2020SkipGNN, ijcai2021p506, zhao2023semignn} have gradually become a research hotspot. Researchers have explored various graph neural network architectures, such as modeling multi-scale features of proteins through hierarchical graph convolution~\cite{gao2023hierarchical}, representing protein microenvironment dependencies using heterogeneous graph~\cite{wu2024mapeppi}, integrating multi-head attention modules with graph attention networks to optimize amino acid sequence expression~\cite{10.1093/bioinformatics/btad052}, and introducing graph isomorphism networks to extract local and global topological features~\cite{Zeng2024GNNGL-PPI}. Huang et al.~\cite{Huang2020SkipGNN} constructed a Skip Graph architecture, extracting multi-layer interaction features by aggregating node information from both second-order neighbors and direct neighbors simultaneously. Kang et al.~\cite{10.1093/bib/bbab513} developed an encoder framework based on GCN, combining link representation learning methods to effectively capture association patterns between molecules. Lv et al.~\cite{ijcai2021p506} designed a PPI evaluation framework, utilizing graph networks to model interaction features between proteins, providing a new perspective for protein function prediction. Baranwal et al.~\cite{Baranwal2022Struct2Graph} transformed the globular structure of folded proteins into graph structures, achieving efficient representation learning of proteins through graph networks. Jha et al.~\cite{Jha2022Prediction} modeled proteins as residue contact networks and combined protein language models to extract residue-level features, providing a new perspective for structure-driven PPI prediction. However, current methods still face three bottlenecks: (1) insufficient capability in extracting residue-level structural features, leading to the loss of important physicochemical properties; (2) high memory consumption due to complex graph computation processes, limiting model scalability; (3) limited generalization performance for new proteins, struggling to handle unknown interaction scenarios. These deficiencies significantly impact the application potential of PPI prediction models in real-world biomedical scenarios.

\section{Methodology}\label{sec:Methodology}
\subsection{Problem statement}
\begin{figure*}[htbp]
\centering
\includegraphics[width=\linewidth]{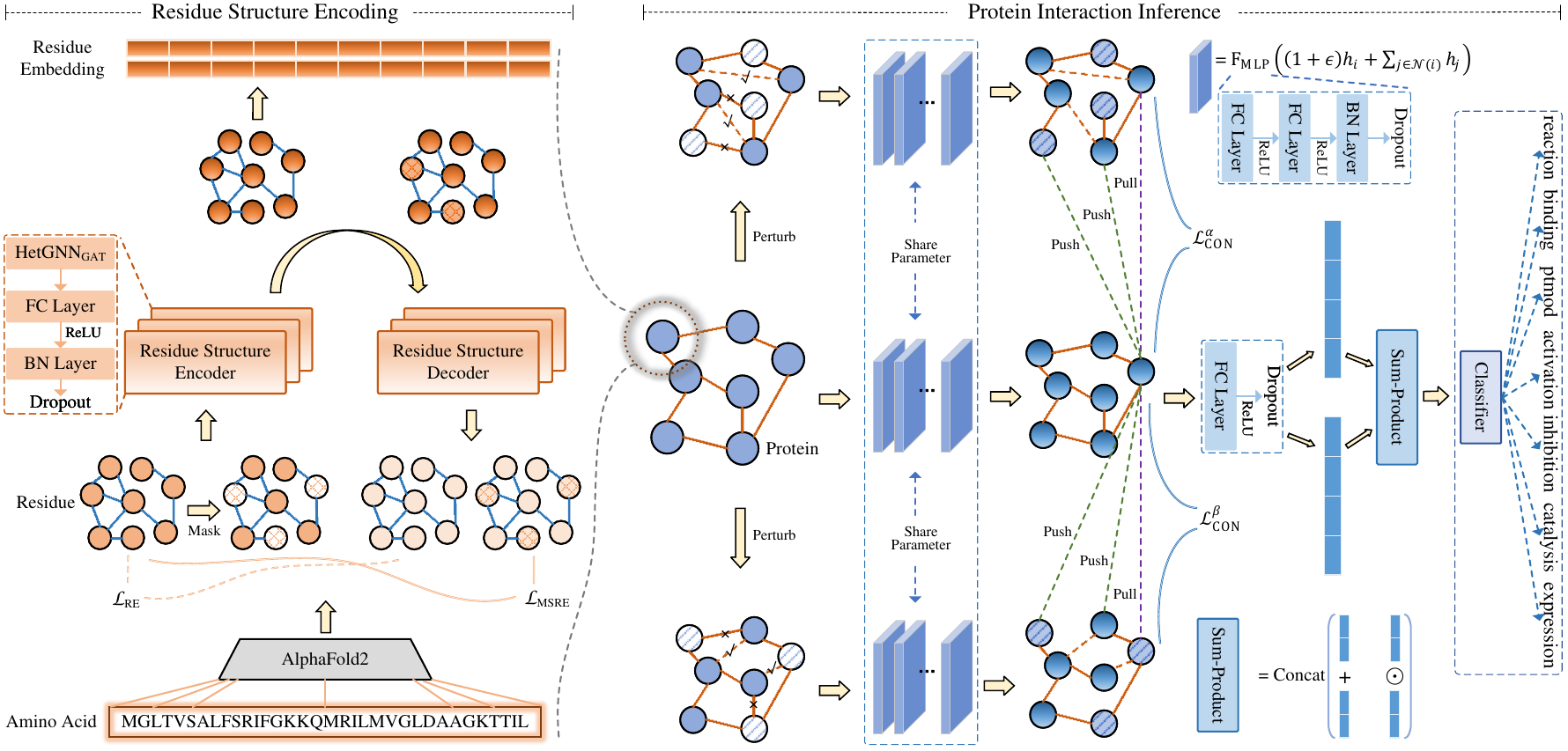}
\caption{Overall workflow of the proposed JmcPPI.}
\label{fig:overall_workflow}
\end{figure*}
The PPI prediction task can be formally defined as follows: Given a pair of proteins \((p_i, p_j)\), we aim to infer the specific type of interaction \(y_{ij}\) between them based on their sequence properties and structural characteristics. Extensive prior research has demonstrated that GNNs, due to their modeling prowess with complex topological relationships, have achieved remarkable advancements in areas such as recommendation systems~\cite{10.1145/3535101}, traffic forecasting~\cite{10077454}, and emotion recognition~\cite{li2024tracing}. In this work, our JmcPPI integrates the topological modeling capabilities of GNNs to synergistically extract both the internal structural features and external interaction patterns of proteins. As depicted in Figure~\ref{fig:overall_workflow}, the proposed approach employs a two-stage progressive architecture, comprising residue structure encoding and protein interaction inference. The former can be regarded as a protein pre-training module, taking the protein structure graph as input; the latter serves as the downstream module for PPI prediction, taking the protein interaction graph as input.

\textit{Protein structure graph:} Given a protein $p$, its structure is composed of a series of amino acid residues, each of which can be characterized by various physicochemical properties. Drawing on the research of Gao et al.~\cite{gao2023hierarchical}, we select seven key features, including topological polar surface area, polarity, isoelectric point, acidity and alkalinity, octanol-water partition coefficient, hydrogen bond donor, and hydrogen bond acceptor, as the node attributes of residues. Concurrently, referring to the work of Zhang et al.~\cite{zhang2023protein}, we define three types of topological relations between residues, including sequential, radial, and K-nearest neighbor relationships, to depict the spatial connection patterns between residues. Based on the aforementioned definitions, the protein structure graph can be represented as a heterogeneous graph $\mathcal{G}_p=(\mathcal{V}_p, \mathcal{E}_p)$. Here, $\mathcal{V}_p$ represents the set of nodes, with node $v_i \in \mathcal{V}_p$ denoting the $i$-th residue in protein $p$; $\mathcal{E}_p$ represents the set of edges, with edge $e_{ij} \in \mathcal{E}_p$ indicating the connection between residues $v_i$ and $v_j$. Furthermore, the edge set $\mathcal{E}_p$ can be decomposed into three subsets, i.e., $\mathcal{E}_p=\mathcal{E}_{ps} \cup \mathcal{E}_{pr} \cup \mathcal{E}_{pk}$. Here, $\mathcal{E}_{ps}$, $\mathcal{E}_{pr}$, and $\mathcal{E}_{pk}$ represent the sequential edges, radial edges, and K-nearest neighbor edges, respectively. Thus, $\mathcal{G}_p$ can be decomposed into three homogeneous graphs $\mathcal{G}_{ps}=(\mathcal{V}_p, \mathcal{E}_{ps})$, $\mathcal{G}_{pr}=(\mathcal{V}_p, \mathcal{E}_{pr})$, and $\mathcal{G}_{pk}=(\mathcal{V}_p, \mathcal{E}_{pk})$.

\textit{Protein interaction graph:} Pairs of proteins constitute numerous interactive relationships, thus forming a PPI network. In this work, we construct a protein interaction graph based on the PPI network, formally denoted as \( \mathcal{G}_q = (\mathcal{N}_q, \mathcal{R}_q) \). Here, \( \mathcal{N}_q \) represents the set of nodes, where node \( n_i \in \mathcal{N}_q \) indicates the \( i \)-th protein in the PPI network; \( \mathcal{R}_q \) represents the set of edges, with edge \( r_{ij} \in \mathcal{R}_q \) signifying the connection between proteins \( n_i \) and \( n_j \). In JmcPPI, the initial representation of a protein is derived by calculating the mean of all its residue features. Based on the protein interaction graph \( \mathcal{G}_q \), the PPI prediction task can be transformed into learning a parameterized function \( \text{F} \), which can infer the corresponding interaction type \( Y \) from \( \mathcal{G}_q \).

\subsection{Residue structure encoding}
We utilize HetGNN~\cite{10.1145/3292500.3330961} as the central component of the residue structure encoder to thoroughly mine the heterogeneous structural information among residues. In the process of aggregating neighbor information, taking into account the potential substantial differences in contributions from various residue nodes to the target residue, we incorporate an aggregation operator based on GAT~\cite{veličković2018graph} within the HetGNN. By harnessing the adaptive weight assignment capability of GAT, the model can automatically learn the significance levels of different residue neighbors, thereby endowing each residue node with distinct aggregation weights.

We employ a fully connected layer to perform dimensionality transformation on residue features:
\begin{equation}
X_{\text{F}} = \text{FC}(X),
\end{equation}
where $\text{FC}$ denotes the fully connected layer. HetGNN based on the GAT operator can be formally represented as:
\begin{equation}
\begin{aligned}
X_{\text{H}} &= \text{HetGNN}_{\text{GAT}}(\mathcal{E}_p, X_{\text{F}}), \\
&= \sum_{\varpi \in \mathcal{T}_{\text{ype}}} \left( \text{GAT}_{\varpi} (\mathcal{E}_{\varpi}, X_{\text{F}})\right),
\end{aligned}
\end{equation}
where $\mathcal{T}_{\text{ype}}$ represents the set of edge types, that is, $\mathcal{T}_{\text{ype}} = \{ps, pr, pk\}$. In order to reduce the model's memory consumption, we adopt the dot-product version of attention coefficients rather than the original concatenation version when calculating attention weights, inspired by the Transformer~\cite{vaswani2017attention}. The dot-product GAT can be expressed as:
\begin{equation}
\begin{split}
& x_{\text{G}, i} = \sum_{j \in {N}(i)} \delta_{ij} W_{\text{G}} x_j, \\
& \text{s.t.}\ \delta_{ij} = \frac{\exp ((W_1 x_i)^\top \odot W_2 x_j)}
     {\sum_{k \in {N}(i)} \exp ((W_1 x_i)^\top \odot W_3 x_k)}.
\end{split}
\end{equation}
Here, $\delta_{ij}$ represents the attention coefficient between residues $v_i$ and $v_j$; $\odot$ is the dot-product (Hadamard product) operation, and ${N}(i)$ denotes the neighbor indices of the $i$-th residue; $W_{\text{G}}$, $W_1$, $W_2$, and $W_3$ are trainable parameters. Following the setup of Veličković et al.~\cite{veličković2018graph}, we utilize multi-head GAT to enhance the model's stability.

To further enhance the feature representation of amino acid residues, we sequentially pass the obtained residue features \(X_{\text{H}}\) through a fully connected layer (FC), a ReLU activation layer, a batch normalization layer (BN), and a dropout layer (DP). This process can be formally represented as:
\begin{equation}
X_{\text{E}} = \text{DP}\left(\text{BN}\left(\text{ReLU}\left(\text{FC}\left(X_{\text{H}}\right)\right)\right)\right).
\end{equation}
By stacking the aforementioned network architecture (residue structure encoder) in multiple layers, we ultimately obtain the encoded residue features \( \mathrm{X}_{\text{E}} \).

We implement a self-supervised learning strategy~\cite{9770382, NEURIPS2019_a2b15837}, which entails the reconstruction of residue features to refine the network parameters of the encoder. Specifically, we employ the symmetric architecture of the encoder as a residue structure decoder, thereby generating reconstructed features corresponding to the original features $X$. The residue decoding process can be formally represented as follows:
\begin{equation}
\begin{split}
&\mathrm{X}_{\text{H}} = \text{HetGNN}_{\text{GAT}}(\mathcal{E}_{p}, \mathrm{X}_{\text{E}}), \\
&\mathrm{X}_{\text{D}} = \text{DP}(\text{BN}(\text{ReLU}(\text{FC}(\mathrm{X}_{\text{H}})))).
\end{split}
\end{equation}
After stacking multiple layers of decoders, we obtain the decoded residue features $\hat{X}_{\text{D}}$, and utilize a fully connected layer to transform its feature dimensions to align with those of the original features $X$:
\begin{equation}
\hat{X} = \text{FC}(\hat{X}_{\text{D}}).
\end{equation}

We define a training objective aimed at optimizing the network parameters by minimizing the distance between the reconstructed features $\hat{X}$ and the original features $X$:
\begin{equation}
\mathcal{L}_{\text{RE}} = \frac{1}{M} \sum_{i=1}^{M} \left( x_{i} - \hat{x}_{i} \right)^2,
\end{equation}
where $M$ represents the number of residues, $x_{i} \in X$ denotes the original representation of the $i$-th residue, and $\hat{x}_{i} \in \hat{X}$ indicates the reconstructed representation of the $i$-th residue.

\subsection{Masked residue reconstruction}
Masked modeling, which leverages the intrinsic structure of data to enhance the model's comprehension and generalization capabilities, has achieved remarkable success in various fields such as visual analysis~\cite{NEURIPS2023_9ed1c94a}, multimodal interaction~\cite{ChenEVE2024}, graph learning~\cite{liu2023rethinking}, language processing~\cite{chen-etal-2023-cheaper}, and bioinformatics~\cite{chen2024pepmlm}. In this study, we introduce a masked residue reconstruction task that shares parameters with the standard reconstruction task, aiming to further augment the expressive power of the residue structure encoder.

Initially, we employ a masking vector $m_\text{s}$ to randomly mask the original features $X$ at a predefined ratio, thereby generating the masked residue features $X_{\text{MS}}$:
\begin{equation}
x_{\text{MS},i} =  
\begin{cases}  
m_\text{s} & \text{if } x_i \in X_{\text{SUB}}, \\  
x_i & \text{if } x_i \notin X_{\text{SUB}}.  
\end{cases}
\end{equation}
Here, $1 \leq i \leq M$, $x_i \in X$, and $x_{\text{MS},i} \in X_{\text{MS}}$; $X_{\text{SUB}}$ is a subset randomly sampled from the original residue features at a specified ratio, designated for the masking operation. The fullly connected layer is used to transform the feature dimensions of $X_{\text{MS}}$:
\begin{equation}
X_{\text{MSF}} = \text{FC}(X_{\text{MS}}).
\end{equation}

Subsequently, $X_{\text{MS}}$ is fed into an residue structure encoder with shared parameters, followed by a residue structure decoder with shared parameters:
\begin{equation}
\begin{split}
&X_{\text{MSH}} = \text{HetGNN}_{\text{GAT}}(\mathcal{E}_p, X_{\text{MSF}}), \\
&X_{\text{MSE}} = \text{DP}(\text{BN}(\text{ReLU}(\text{FC}(X_{\text{MSH}})))), \\
&\mathrm{X}_{\text{MSH}} = \text{HetGNN}_{\text{GAT}}(\mathcal{E}_p, \mathrm{X}_{\text{MSE}}), \\
&\mathrm{X}_{\text{MSD}} = \text{DP}(\text{BN}(\text{ReLU}(\text{FC}(\mathrm{X}_{\text{MSH}})))). \\
\end{split}
\end{equation}
In this process, $\mathrm{X}_{\text{MSE}}$ is the masked encoded features after multi-layer stacking. By stacking multiple layers of residue structure decoders, the masked decoded feature $\hat{X}_{\text{MSD}}$ is obtained. To attain the masked reconstructed features $\hat{X}_{\text{MS}}$ with the same dimensions as $X$, we apply a fully connected layer to $\hat{X}_{\text{MSD}}$ for dimension transformation:
\begin{equation}
\hat{X}_{\text{MS}} = \text{FC}(\hat{X}_{\text{MSD}}).
\end{equation}

We define a masked reconstruction loss \( L_{\text{MSRE}} \), serving as the training objective for the masked residue reconstruction task, to reduce the distance between \( \hat{X}_{\text{MS}} \) and the original features \( X \):
\begin{equation}
\mathcal{L}_{\text{MSRE}} = \frac{1}{M} \sum_{i=1}^{M} \left(1 - \frac{x_{i}^\top \hat{x}_{\text{MS},i}}{\Vert x_{i} \Vert \times \Vert \hat{x}_{\text{MS},i} \Vert}\right)^\delta.
\end{equation}
Here, \( \delta \) is a scaling factor used to adjust the weight of samples; \( \hat{x}_{\text{MS},i} \in \hat{X}_{\text{MS}} \) represents the masked reconstructed representation of the \( i \)-th residue. Based on this, the overall objective function for the residue structure encoding phase is defined as:
\begin{equation}
\mathcal{L}_{\text{STR}} = \mathcal{L}_{\text{RE}} + \gamma_{\text{STR}} \mathcal{L}_{\text{MSRE}} + \lambda_{\text{STR}} \lVert W_{\text{STR}} \rVert,
\end{equation}
Here, \( \gamma_{\text{STR}} \) is the weighting factor used to balance the standard reconstruction loss \( L_{\text{RE}} \) with the masked reconstruction loss \( L_{\text{MSRE}} \); \( \lambda_{\text{STR}} \) is the regularization coefficient used to control model complexity; \( W_{\text{STR}} \) represents the trainable parameters of the first phase.

Finally, for the \(i\)-th protein \(n_i\), we take the mean of the output \( \mathrm{X}_{\text{E},i} \) from the residue structure encoder to obtain the representation of protein \(n_i\):
\begin{equation}
h_i = \frac{1}{M} \sum_{j=1}^{M} \mathrm{x}_{\text{E},i,j}.
\end{equation}
Here, \(M\) denotes the number of residues in protein \(n_i\), \( \mathrm{x}_{\text{E},i,j} \in \mathrm{X}_{\text{E},i} \) denotes the feature of the \(j\)-th residue in the \(i\)-th protein, and \(h_i \in H\) is the representation of the \(i\)-th protein.

\subsection{Protein interaction inference}
To capture the extrinsic interaction information between proteins, we employ GIN~\cite{xu2018how} as the protein interaction encoder and take the constructed protein interaction graph $\mathcal{G}_q$ as input. In GIN, neighbor information is aggregated through the summation operation and the multi-layer perceptron (MLP) is used as the update function. The protein interaction encoder can be formally represented as:
\begin{equation}
\begin{aligned}
&H_{\text{E}} = \text{GIN}_{\text{MLP}}(\mathcal{R}_q, H), \\
&\text{i.e.,} \ h_{\text{E},i} = \text{F}_{\text{MLP}} \left( (1 + \epsilon) h_i + \sum_{j \in {N}(i)} h_j \right),
\end{aligned}
\end{equation}
where $h_{\text{E},i} \in H_{\text{E}}$ denotes the encoded feature of protein $n_i$, and $\epsilon$ is a learnable parameter; $h_j \in H$ represents the representation of the neighboring protein $n_j$ of $h_i$; $\text{F}_{\text{MLP}}$ denotes the MLP function, the structure of which is as follows:
\begin{equation}
\text{F}_{\text{MLP}}(\cdot) = \text{DP}(\text{BN}(\text{ReLU}(\text{FC}(\text{ReLU}(\text{FC}(\cdot)))))).
\end{equation}

After stacking multiple layers of encoders, we obtain the encoded protein features $\mathrm{H}_{\text{E}}$. Subsequently, these features are passed through a fully connected layer (FC), ReLU activation layer, and dropout layer (DP):
\begin{equation}
\hat{H} = \text{DP}(\text{ReLU}(\text{FC}(\mathrm{H}_{\text{E}}))).
\end{equation}
Referring to the Sum-Product function designed by Li et al.~\cite{10081075}, we simultaneously employ vector addition and multiplication operations to integrate the features of proteins \( n_i \) and \( n_j \) to represent their interactions:
\begin{equation}
\hat{y}_{ij} = \text{FC}(\text{CAT}(\hat{h}_i + \hat{h}_j, \hat{h}_i \odot \hat{h}_j)).
\end{equation}
Here, \( \odot \) denotes the Hadamard product (element-wise multiplication) operation, and \( \text{CAT} \) represents the feature concatenation operation.

Given the training set \( \mathcal{D}_{\text{Train}} = (\mathcal{X}, \mathcal{Y}) \), we adopt the multi-class cross-entropy as the training objective to optimize the protein encoder. The loss function for this part can be formally represented as:
\begin{equation}
\mathcal{L}_{\text{IN}} = -\frac{1}{|\mathcal{X}|} \sum_{(n_i,n_j) \in \mathcal{X}} \sum_{c=1}^{C} \left( y_{ij}^c \log (\hat{y}_{ij}^c) + (1-y_{ij}^c) \log (1-\hat{y}_{ij}^c) \right).
\end{equation}
Here, \( \mathcal{X} \) denotes the set of protein pairs in the training set, and \( |\mathcal{X}| \) represents the number of protein pairs; \( \mathcal{Y} \) is the corresponding ground-truth PPI type labels, with \( y_{ij} \in \mathcal{Y} \); \( C \) represents the total number of interaction types, and \( c \in C \) is the index of a specific interaction type.

\subsection{Multi-graph contrastive learning}
Contrastive learning, by comparing the feature representations of positive and negative samples in the latent space, effectively extracts information about similarity and dissimilarity within the data, and has achieved significant success in fields such as computer vision~\cite{9873966}, natural language processing~\cite{10731549}, and recommender systems~\cite{10872817}. However, existing PPI prediction methods tend to predict interactions between seen proteins and perform poorly when faced with novel proteins. To address this issue, we construct a multi-graph contrastive learning task aimed at maximizing the feature consistency of the same protein across different graphs while minimizing the feature similarity between different proteins. This design enables the protein interaction encoder to extract potential universal representations of proteins from limited information, reducing reliance on seen proteins during the message passing process, and thereby better uncovering interaction cues between novel proteins. We perturb the original protein interaction graph $\mathcal{G}_q$ with a certain probability to generate two distinct new graphs $\mathcal{G}_q^\alpha$ and $\mathcal{G}_q^\beta$. Taking $\mathcal{G}_q^\alpha$ as an example, we describe the perturbation process from both the node and edge dimensions.

\textit{Node perturbation:} Given a node perturbation probability $\rho^\alpha$, we generate a node perturbation matrix $B^\alpha$ and perform element-wise multiplication with the original node features $H$, setting some elements of $H$ to 0 to achieve node perturbation. Mathematically, node perturbation can be represented as:
\begin{equation}
H^\alpha = H \odot B^\alpha,
\end{equation}
where $H^\alpha$ denotes the perturbed protein representations; $B^\alpha$ has the same shape as $H$, and $b_{ij}^\alpha$ represents the elements of $B^\alpha$; $b_{ij}^\alpha \in \{0,1\}$ takes the value 0 with probability $\rho^\alpha$ and the value 1 with probability $1-\rho^\alpha$.

\textit{Edge perturbation:} Given an edge perturbation probability $\rho^\alpha$, we generate an edge perturbation vector $O^\alpha$ and reassign new source and target nodes to each element, thereby creating new edges. Mathematically, edge perturbation can be represented as:
\begin{equation}
\begin{aligned}
&\mathcal{S}_i^\alpha =  
\begin{cases}  
\text{SA}(0,|\mathcal{N}_q|-1) & \text{if } i \in O^\alpha, \\  
\mathcal{S}_i & \text{if } i \notin O^\alpha,  
\end{cases} \\
&\mathcal{T}_i^\alpha =  
\begin{cases}  
\text{SA}(0,|\mathcal{N}_q|-1) & \text{if } i \in O^\alpha, \\  
\mathcal{T}_i & \text{if } i \notin O^\alpha.  
\end{cases}
\end{aligned}
\end{equation}
Here, $\text{SA}(0,|\mathcal{N}_q|-1)$ denotes random sampling from the node indices, and $|\mathcal{N}_q|$ represents the number of nodes in graph $\mathcal{G}_q$; $O^\alpha$ is the edge perturbation vector and represents the indices of edges to be perturbed, with length $|O^\alpha| = \rho^\alpha |R_q|$; $\mathcal{S}$ and $\mathcal{T}$ represent the source and target node indices of edges in graph $\mathcal{G}_q$, and $R_q = (\mathcal{S}, \mathcal{T})$; $\mathcal{S}^\alpha$ and $\mathcal{T}^\alpha$ represent the perturbed source and target node indices, $R_q^\alpha = (\mathcal{S}^\alpha, \mathcal{T}^\alpha)$, and $R_q^\alpha$ denotes the set of perturbed edges.

After node perturbation and edge perturbation, we can obtain a new PPI graph $\mathcal{G}_q^\alpha=(\mathcal{N}_q^\alpha, \mathcal{R}_q^\alpha)$. Similarly, after similar perturbations, another new PPI graph $\mathcal{G}_q^\beta=(\mathcal{N}_q^\beta, \mathcal{R}_q^\beta)$ can be derived. Next, these two graphs are input into protein interaction encoders with shared parameters:
\begin{equation}
\begin{aligned}
&H_{\text{E}}^\alpha = \text{GIN}_{\text{MLP}}(\mathcal{R}_q^\alpha, H^\alpha), \\
&H_{\text{E}}^\beta = \text{GIN}_{\text{MLP}}(\mathcal{R}_q^\beta, H^\beta).
\end{aligned}
\end{equation}
After stacking multiple layers of encoders, the encoded protein features $\mathrm{H}_{\text{E}}^\alpha$ and $\mathrm{H}_{\text{E}}^\beta$ are obtained respectively. For $\mathrm{H}_{\text{E}}$ and $\mathrm{H}_{\text{E}}^\alpha$, we construct a contrastive loss $L_{\text{CON}}^\alpha$ to pull closer the distances of the same protein features and push away the distances of different protein features:
\begin{equation}
\mathcal{L}_{\text{CON}}^\alpha = -\frac{1}{|\mathrm{H}_{\text{E}}|} \sum_{i=1}^{|\mathrm{H}_{\text{E}}|} \log \left( \frac{\exp \left( \frac{\mathrm{h}_{\text{E},i} \cdot \mathrm{h}_{\text{E},i}^\alpha}{\tau} \right)}{\sum_{j \neq i} \exp \left( \frac{\mathrm{h}_{\text{E},i} \cdot \mathrm{h}_{\text{E},j}^\alpha}{\tau} \right)} \right).
\end{equation}
Here, $|\mathrm{H}_{\text{E}}^\alpha|$ denotes the number of proteins, and $\tau$ is the temperature coefficient; $\mathrm{h}_{{\text{E}},i}^\alpha \in \mathrm{H}_{\text{E}}^\alpha$ represents the positive sample feature, that is, the perturbed feature of protein $n_i$ corresponding to $\mathrm{h}_{\text{E},i}$. Similarly, we construct the contrastive loss $L_{\text{CON}}^\beta$ for protein features $\mathrm{H}_{\text{E}}$ and $\mathrm{H}_{\text{E}}^\beta$:
\begin{equation}
\mathcal{L}_{\text{CON}}^\beta = -\frac{1}{|\mathrm{H}_{\text{E}}|} \sum_{i=1}^{|\mathrm{H}_{\text{E}}|} \log \left( \frac{\exp \left( \frac{\mathrm{h}_{\text{E},i} \cdot \mathrm{h}_{\text{E},i}^\beta}{\tau} \right)}{\sum_{j \neq i} \exp \left( \frac{\mathrm{h}_{\text{E},i} \cdot \mathrm{h}_{\text{E},j}^\beta}{\tau} \right)} \right).
\end{equation}
Combining the above two contrastive losses, we obtain the complete loss function $L_{\text{CON}}$ for the multi-graph contrastive learning task:
\begin{equation}
\mathcal{L}_{\text{CON}} = \mathcal{L}_{\text{CON}}^\alpha + \mathcal{L}_{\text{CON}}^\beta.
\end{equation}

In the proposed JmcPPI, the adoption of a multi-graph contrastive learning strategy offers the following advantages: (1) Learning of general features -- By learning general feature representations of proteins from limited information in an unsupervised manner, the model reduces its reliance on known proteins in the training set, thereby enhancing its generalization ability on novel proteins. (2) Enhancing robustness -- Through multi-perspective learning, such as node and edge perturbations, the protein encoder's adaptability to different interaction patterns and protein variants is strengthened, improving the model's robustness and flexibility. (3) Capturing latent relationships -- Via employing the contrastive learning task, the model can effectively capture latent relationships between proteins, such as differences and similarities, thus better generalizing to unseen proteins and their interaction scenarios.

Finally, the overall loss for the protein interaction inference phase with multi-graph contrastive learning is as follows:
\begin{equation}
\mathcal{L}_{\text{IN-CON}} = \mathcal{L}_{\text{IN}} + \gamma_{\text{IN-CON}} \mathcal{L}_{\text{CON}} + \lambda_{\text{IN-CON}} \lVert W_{\text{IN-CON}} \rVert,
\end{equation}
where \( \gamma_{\text{IN-CON}} \) is the weighting parameter, \( \lambda_{\text{IN-CON}} \) is the regularization factor, and \( W_{\text{IN-CON}} \) represents the learnable parameters of the second phase.

\section{Experimental Setup}\label{sec:Experimental_Setup}
\subsection{Benchmark dataset}
Building upon prior studies~\cite{ijcai2021p506,zhao2023semignn,wu2024mapeppi}, we evaluate JmcPPI's performance on three widely-used benchmark datasets, i.e., SHS27k, SHS148k, and STRING. These datasets collectively cover seven protein interaction types, including reaction, binding, post-translational modifications (ptmod), activation, inhibition, catalysis, and expression. The STRING dataset, derived from the STRING database~\cite{10.1093/nar/gky1131}, contains 1,150,830 Homo sapiens PPI entries involving 14,952 proteins and 572,568 interactions. SHS27k and SHS148k are STRING subsets comprising 16,912 and 99,782 PPI entries with 7,401 and 43,397 interaction pairs, respectively. In these datasets, each pair of proteins may correspond to multiple interactions, so PPI prediction is a multi-label classification task. To control sequence homology bias, we follow Wu et al.~\cite{wu2024mapeppi} to filter proteins with amino acid length $\geq$50 and sequence homology $<$40\% for subset construction. To mitigate overfitting risks from training-test set overlap, we adopt breadth-first search (BFS), depth-first search (DFS), and random partition strategies based on existing works~\cite{ijcai2021p506,wu2024mapeppi}. The datasets are split into training, validation, and test sets at a 3:1:1 ratio, and detailed data statistics are provided in Table~\ref{tab:dataset}.
\begin{table}[htbp]
\centering
\scriptsize
\setlength{\tabcolsep}{9pt}
\caption{\label{tab:dataset}Statistical information of the adopted benchmark datasets.}
% \begin{threeparttable}
\begin{tabular}{ccccc}
\toprule
Dataset & Scheme & Training & Validation & Test \\
\midrule
\multirow{3}{*}{SHS27k} & Random &4,440 &1,480 & 1,481 \\
& BFS &4,427 &1,480 &1,494 \\
& DFS &4,427 &1,480 &1,494 \\
\midrule
\multirow{3}{*}{SHS148k} & Random &26,038 &8,679 &8,680 \\
& BFS & 26,000 & 8,679 & 8,718 \\
& DFS & 26,018 & 8,679 & 8,700 \\
\midrule
\multirow{3}{*}{STRING} & Random & 343,540 & 114,514 & 114,514 \\
& BFS & 343,529 & 114,513 & 114,526 \\
& DFS & 342,555 & 114,513 & 115,500 \\
\bottomrule
\end{tabular}
\end{table}

\subsection{Comparison baseline}
In this study, we integrate multiple baseline methods to validate the superiority of JmcPPI. Based on protein modeling strategies, existing approaches are categorized into two classes: sequence-driven methods and structure-driven methods. Sequence-driven methods predict interactions by analyzing amino acid sequences, with representative models including DPPI~\cite{10.1093/bioinformatics/bty573}, DNN-PPI~\cite{molecules23081923}, PIPR~\cite{10.1093/bioinformatics/btz328}, GNN-PPI~\cite{ijcai2021p506}, and SemiGNN-PPI~\cite{zhao2023semignn}. Structure-driven methods rely on three-dimensional conformations or microenvironmental features of proteins, exemplified by HIGH-PPI~\cite{gao2023hierarchical} and MAPE-PPI~\cite{wu2024mapeppi}.

DPPI constructed predictive models using protein sequence data, leveraging deep learning to efficiently process large-scale training data and captured complex interaction patterns. DNN-PPI employed a deep neural network architecture, where raw amino acid sequences were processed through encoding, embedding, CNN, and LSTM layers to autonomously learn features from primary structures for interaction prediction. PIPR utilized an end-to-end residual recurrent CNN to integrate local features with global contextual information, enhancing prediction accuracy. GNN-PPI innovatively adopted breadth-first and depth-first search strategies to construct test sets, optimizing data partition for novel protein interactions. SemiGNN-PPI combined GNN with the mean teacher framework through multi-graph construction and label dependency analysis, effectively leveraging unlabeled graph-structured PPI data. HIGH-PPI proposed a dual-view hierarchical model, the bottom view constructed protein graphs with amino acid residues as nodes and physical adjacency as edges, while the top view built PPI graphs using protein graphs as nodes and interactions as edges, employing temporal graph neural networks to learn relational features. MAPE-PPI introduced a masked codebook pre-training strategy by sensing chemical and geometric properties of amino acid residues, capturing dependencies across diverse microenvironments.

\subsection{Implementation detail}
All experiments are conducted on a server equipped with a single NVIDIA GeForce RTX 3090 GPU, utilizing Ubuntu 20.04 operating system and PyTorch 2.4.1 framework. We employ AlphaFold2~\cite{Jumper2021Highly} for 3D protein structure extraction from amino acid sequences. The model is trained with an Adam optimizer~\cite{KingmaB14Adam}, using an L2 regularization coefficient of 1e-4 and an initial learning rate of 1e-3. Key architectural parameters include: (1) For residue structure encoding, both encoder and decoder contain 4-layer networks with 5-head GAT, hidden dimension of 128, batch size of 128, and maximum 50 training epochs; (2) For protein interaction representation, the encoder adopts 3-layer networks with expanded hidden dimension of 1024 and maximum 800 training epochs. Other hyperparameter configurations are provided in Table~\ref{tab:hyperparameter}.
\begin{table}[htbp]
\centering
\scriptsize
\setlength{\tabcolsep}{5pt}
\caption{\label{tab:hyperparameter}Setting details for partial hyperparameters. DR denotes the dropout rate, MR denotes the mask rate, SF denotes the scale factor, PR denotes the perturbation rate, and TC denotes the temperature coefficient; $\gamma_{\text{STR}}$ and $\gamma_{\text{IN-CON}}$ serve as trade-off factors for losses in the two stages.}
% \begin{threeparttable}
\begin{tabular}{ccccccccc}
\toprule
Dataset & Scheme & DR & MR & SF & PR & TC & $\gamma_{\text{STR}}$ & $\gamma_{\text{IN-CON}}$ \\
\midrule
\multirow{3}{*}{SHS27k} & Random & 0.2 & 0.25 & 1.5 & 0.1 & 0.6 & 0.5 &0.6 \\
& BFS &0.2 &0.25 &1.5 &0.1 &0.1 & 0.5 &0.5 \\
& DFS &0.2 &0.25 &1.5 &0.25 &0.2 & 0.5 &0.6 \\
\midrule
\multirow{3}{*}{SHS148k} & Random &0.3 &0.25 &1.5 &0.1 &0.4 & 0.5 &0.6 \\
& BFS &0.2 &0.25 &1.5 &0.25 &0.2 & 0.5 &0.5 \\
& DFS &0.2 &0.25 &1.5 &0.25 &0.2 & 0.5 &0.5 \\
\midrule
\multirow{3}{*}{STRING} & Random & 0.1 & 0.25 & 1.5 & 0.1 & 0.2 & 0.5 &0.5 \\
& BFS & 0.2 & 0.25 & 1.5 & 0.25 & 0.2 & 0.5 &0.5 \\
& DFS & 0.2 & 0.25 & 1.5 & 0.25 & 0.2 & 0.5 &0.5 \\
\bottomrule
\end{tabular}
\end{table}

\section{Experimental Result}\label{sec:Experimental_Result}
\subsection{Overall comparison}
\begin{table*}[htbp]
\centering
\scriptsize
\setlength{\tabcolsep}{5.5pt}
\caption{\label{tab:overall_comparison}Overall comparison on public benchmark datasets. The scores (\%) of all baseline methods are obtained from the results of MAPE-PPI~\cite{wu2024mapeppi}, where shadow and underline denote the best and second-best performance, respectively.}
% \begin{threeparttable}
\begin{tabular}{cccccccccccccc}
\toprule
\multirow{2}{*}{Method}  & \multicolumn{4}{c}{SHS27k} &\multicolumn{4}{c}{SHS148k} &\multicolumn{4}{c}{STRING} & \multirow{2}{*}{Average}\\
\cmidrule(lr){2-5}  \cmidrule(lr){6-9} \cmidrule(lr){10-13}
 & Random & BFS & DFS & Average & Random & BFS & DFS & Average & Random & BFS & DFS & Average & \\
\midrule
DPPI~\cite{10.1093/bioinformatics/bty573} &70.45 &43.87 & 43.69 &52.67 &76.10 &50.80 & 51.43 &59.44 &92.49 &54.41 &63.41 &70.10 &60.74 \\
DNN-PPI~\cite{molecules23081923} &75.18 &51.59 &48.90 & 58.56 &85.44 &54.56 & 56.70 &65.57 &81.91 &51.53 &61.34 &64.93 &63.02 \\
PIPR~\cite{10.1093/bioinformatics/btz328} & 79.59 &47.13 &52.19 & 59.64 &88.81 &58.57 & 61.38 &69.59 &93.68 &53.80 &64.97 &70.82 &66.68 \\
GNN-PPI~\cite{ijcai2021p506} &83.65 &63.08 &66.52 &71.08 & 90.87 &69.53 &75.34 &78.58 &94.53 &75.69 &84.28 &84.83 &78.17 \\
SemiGNN-PPI~\cite{zhao2023semignn} &85.57 &67.94 &69.25 &74.25 & 91.40 &71.06 &77.62 &80.03 &94.80 &77.10 &84.85 &85.58 &79.95 \\
HIGH-PPI~\cite{gao2023hierarchical} &86.23 &68.40 &70.24 &74.96 & 91.26 &72.87 &78.18 &80.77 &- &- &- &- &- \\
MAPE-PPI~\cite{wu2024mapeppi} &\cellcolor[gray]{0.8}{88.91} &\underline{70.38} &\underline{71.98} &\underline{77.09} & \underline{92.38} &\underline{74.76} &\underline{79.45} &\underline{82.20} &\underline{96.12} &\underline{78.26} &\underline{86.50} &\underline{86.96} &\underline{82.08} \\
\midrule
JmcPPI (Ours) & \underline{88.80} & \cellcolor[gray]{0.8}{79.65} & \cellcolor[gray]{0.8}{73.51} &\cellcolor[gray]{0.8}{80.65} & \cellcolor[gray]{0.8}{93.00} & \cellcolor[gray]{0.8}{77.66} & \cellcolor[gray]{0.8}{83.29} &\cellcolor[gray]{0.8}{84.65} &\cellcolor[gray]{0.8}{96.94} &\cellcolor[gray]{0.8}{81.53} &\cellcolor[gray]{0.8}{90.59} &\cellcolor[gray]{0.8}{89.69} &\cellcolor[gray]{0.8}{85.00} \\
\bottomrule
\end{tabular}
\end{table*}
Table~\ref{tab:overall_comparison} compares the micro-F1 scores of the proposed JmcPPI with baseline methods. Experimental results indicate that among baseline models, MAPE-PPI achieves the best F1 scores across all benchmark datasets, validating its rationality as a strong baseline. From the model architecture perspective, GNN-based methods significantly outperform traditional deep models (e.g., CNN- and RNN-based approaches), highlighting GNN's advantages in modeling complex topological dependencies between proteins. Notably, all sequence-driven methods underperform structure-driven methods (e.g., HIGH-PPI, MAPE-PPI), suggesting that internal structural information enhances the informative representations and provides critical prior knowledge for interaction pattern mining.

Specifically, our JmcPPI achieves average F1 score improvements of 3.56\%, 2.45\%, and 2.73\% over MAPE-PPI on three public datasets (SHS27k, SHS148k, and STRING), with an overall performance gain of 2.92\%. Regarding data partition strategies, JmcPPI performs comparably to MAPE-PPI under the Random partition for SHS27k but exhibits F1 score improvements of 9.27\% and 1.53\% under the more challenging BFS and DFS partitions, respectively. For SHS148k and STRING datasets, JmcPPI consistently outperforms MAPE-PPI across all partition schemes (Random, BFS, and DFS). For instance, under the BFS partition, JmcPPI achieves F1 score gains of 2.90\% (SHS148k) and 3.27\% (STRING) over MAPE-PPI, while under the DFS partition, the improvements further increase to 3.82\% (SHS148k) and 4.09\% (STRING). These results confirm that JmcPPI effectively addresses the interaction prediction for unseen protein pairs by integrating internal structural features with external interaction patterns, significantly enhancing predictive capability.

\begin{figure*}[htbp]
\centering
{\includegraphics[width=0.33\linewidth]{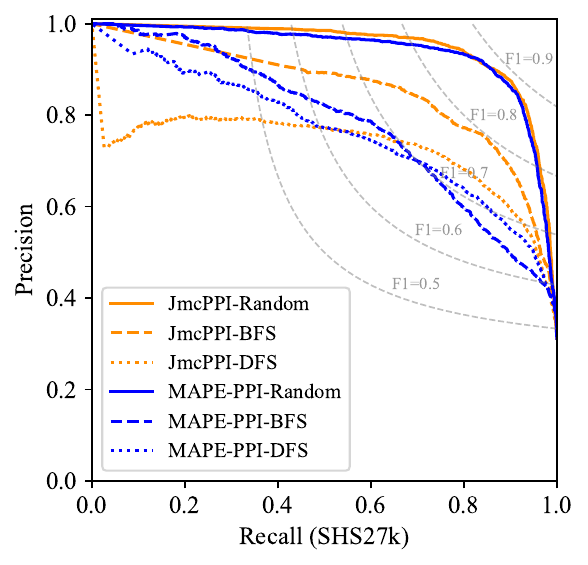}%
\label{fig:AUPR_SHS27k}}
\hfil
{\includegraphics[width=0.33\linewidth]{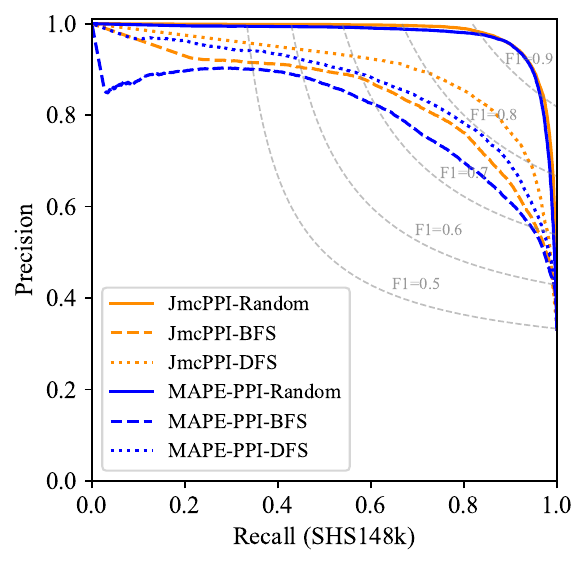}%
\label{fig:AUPR_SHS148k}}
\hfil
{\includegraphics[width=0.33\linewidth]{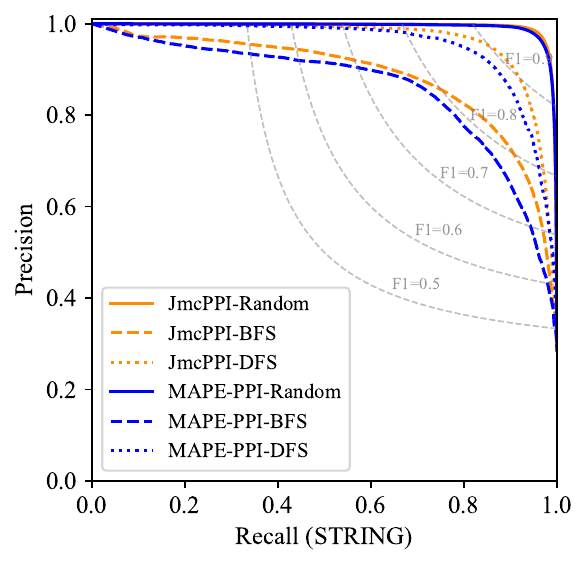}%
\label{fig:AUPR_STRING}}
\caption{\label{fig:AUPR}Precision-recall (PR) curves of JmcPPI and MAPE-PPI on benchmark datasets.}
\end{figure*}
As shown in Figure~\ref{fig:AUPR}, we plot the precision-recall (PR) curves of JmcPPI and MAPE-PPI on three benchmark datasets (SHS27k, SHS148k, and STRING). Under the Random partition, both methods exhibit highly consistent PR curves across all datasets. However, JmcPPI demonstrates substantial advantages under the more challenging BFS and DFS partitions: on SHS27k and SHS148k datasets, its PR curves consistently surpass MAPE-PPI in high-recall regions (recall $>$ 70\%), indicating that JmcPPI maintains higher prediction accuracy even when covering extensive potential protein pairs. For the STRING dataset, the PR curves of JmcPPI as a whole are located above those of MAPE-PPI under BFS and DFS partitions, confirming that JmcPPI still maintains the advanced performance in large-scale scenarios. Additionally, compared to SHS27k and SHS148k, the PR curves of all methods on STRING are closer to the upper-right corner of the coordinate system, implying that large-scale data provides richer interaction patterns for model learning, thereby comprehensively improving prediction performance.

\subsection{Result for each interaction}
\begin{table*}[htbp]
\centering
\scriptsize
\setlength{\tabcolsep}{2.3pt}
\caption{\label{tab:each_PPI}Accuracy and F1 scores for each PPI type on the three datasets.}
% \begin{threeparttable}
\begin{tabular}{cccccccccccccccc}
\toprule
\multirow{2}{*}{Dataset} & \multirow{2}{*}{Scheme} & \multicolumn{7}{c}{Accuracy} & \multicolumn{7}{c}{F1} \\
\cmidrule(lr){3-9}  \cmidrule(lr){10-16}
& & reaction & binding & ptmod & activation & inhibition & catalysis & expression & reaction & binding & ptmod & activation & inhibition & catalysis & expression \\
\midrule
\multirow{3}{*}{SHS27k} & Random & 95.77 & 92.54 & 88.16 & 86.05 & 90.11 & 96.15 & 45.60 & 92.35 & 90.49 & 88.71 & 85.65 & 89.94 & 91.97 & 53.27 \\
& BFS &90.90 & 83.82 & 80.14 & 73.00 & 76.70 & 91.07 & 56.00 & 85.64 & 82.52 & 69.59 & 70.49 & 71.93 & 86.69 & 49.78 \\
& DFS & 79.63 & 80.55 & 82.01 & 77.16 & 77.35 & 77.58 & 34.43 & 78.84 & 82.23 & 72.61 & 56.25 & 67.79 & 76.86 & 37.84 \\
\midrule
\multirow{3}{*}{SHS148k} & Random &96.24 & 94.40 & 91.24 & 91.26 & 94.12 & 96.86 & 54.21 &95.51 & 94.06 & 92.34 & 90.96 & 94.50 & 95.76 & 59.50 \\
& BFS &70.02 & 73.68 & 73.87 & 85.06 & 71.48 & 85.27 & 41.69 &72.58 & 77.66 & 76.10 & 79.58 & 74.57 & 83.69 & 47.76 \\
& DFS &91.40 & 82.49 & 90.88 & 88.88 & 86.51 & 91.53 & 50.96 &86.42 & 83.60 & 81.99 & 76.36 & 83.38 & 87.00 & 53.42 \\
\midrule
\multirow{3}{*}{STRING} & Random &99.14 & 97.61 & 89.45 & 90.64 & 94.97 & 99.10 & 39.49 & 99.14 & 97.61 & 89.45 & 90.64 & 94.97 & 99.10 & 39.49 \\
& BFS & 78.05 & 94.20 & 50.95 & 47.13 & 44.79 & 65.87 & 16.08 & 83.27 & 91.49 & 55.12 & 55.39 & 58.40 & 71.71 & 23.83 \\
& DFS & 92.70 & 94.45 & 77.89 & 76.44 & 82.14 & 91.72 & 32.30 & 92.85 & 93.41 & 83.46 & 81.15 & 86.77 & 92.15 & 40.13 \\
\bottomrule
\end{tabular}
\end{table*}
To investigate the classification performance of different interaction types, Table~\ref{tab:each_PPI} provides fine-grained results of accuracy and F1 scores for each category. Experimental findings reveal that the \textit{expression} category exhibits significant recognition challenges across all datasets. In the SHS27k dataset, the accuracy of \textit{expression} under Random, BFS, and DFS partition schemes is 45.60\%, 56.00\%, and 34.43\%, respectively, markedly lower than other PPI types. Further analysis suggests that the phenomenon may stem from the underrepresentation of \textit{expression} in the training set leads to model bias toward majority classes due to class imbalance (e.g., a share of 4.35\% under the BFS partition and a share of 5.12\% under the DFS partition).

In the SHS148k dataset, the accuracy of \textit{expression} further validates the above conclusions, achieving the lowest values under Random (54.21\%), BFS (41.69\%), and DFS (50.96\%) partition. The performance troughs under BFS and DFS partitions can be attributed to their strict exclusion of seen protein pairs, blocking the model from leveraging historical interaction patterns. For the STRING dataset, \textit{expression} also yielded the lowest accuracy under the three partitions, 39.49\%, 16.08\% and 32.30\%, respectively. Under the BFS partition, in addition to \textit{expression}, \textit{ptmod}, \textit{activation}, and \textit{inhibition} also achieve smaller accuracy rates. Inspection of the data distribution of STRING reveals that they also belong to the minority category as they account for 3.98\%, 9.91\% and 3.92\% of the overall PPI types, respectively. 

\begin{figure*}[htbp]
\centering
{\includegraphics[width=0.33\linewidth]{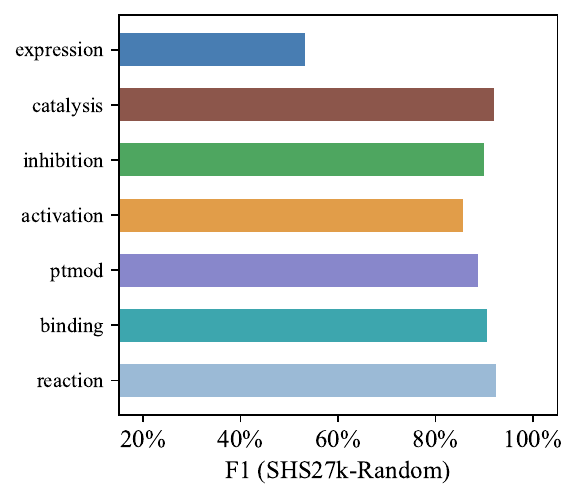}%
\label{fig:each_PPI_SHS27k-Random}}
\hfil
{\includegraphics[width=0.33\linewidth]{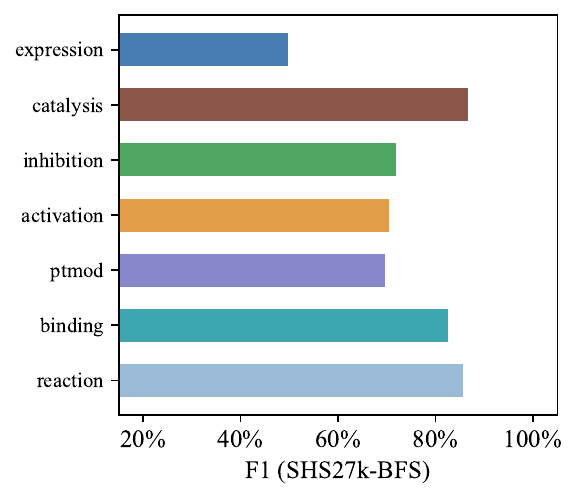}%
\label{fig:each_PPI_SHS27k-BFS}}
\hfil
{\includegraphics[width=0.33\linewidth]{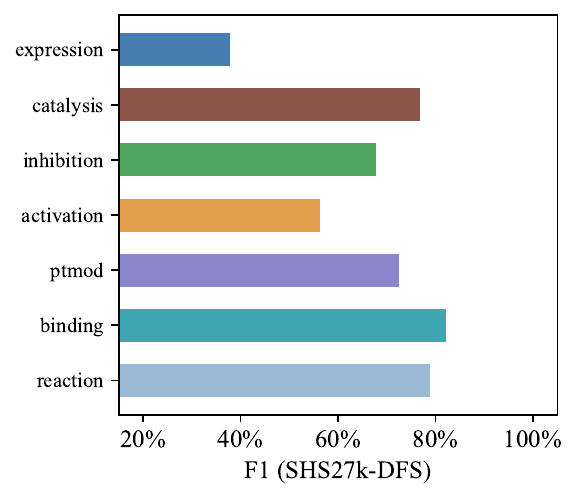}%
\label{fig:each_PPI_SHS27k-DFS}}
\vfil
{\includegraphics[width=0.33\linewidth]{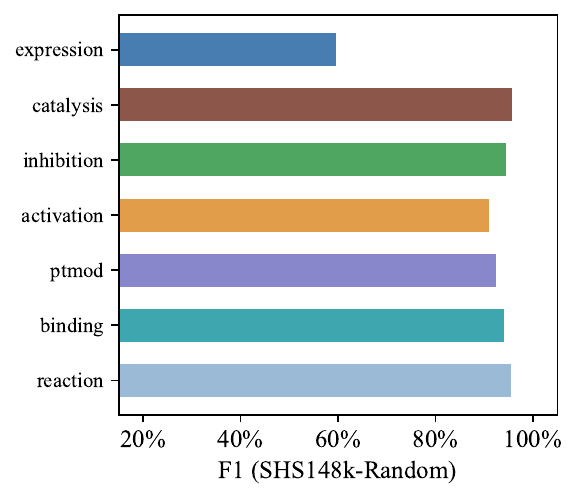}%
\label{fig:each_PPI_SHS148k-Random}}
\hfil
{\includegraphics[width=0.33\linewidth]{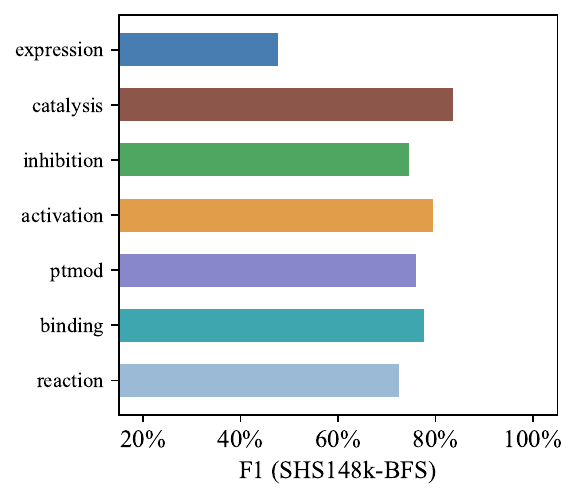}%
\label{fig:each_PPI_SHS148k-BFS}}
\hfil
{\includegraphics[width=0.33\linewidth]{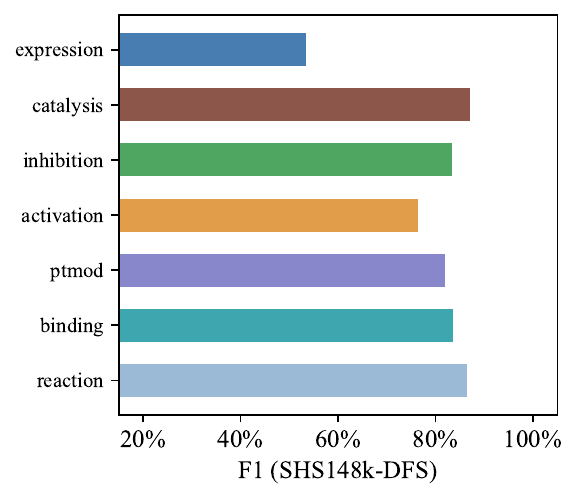}%
\label{fig:each_PPI_SHS148k-DFS}}
\vfil
{\includegraphics[width=0.33\linewidth]{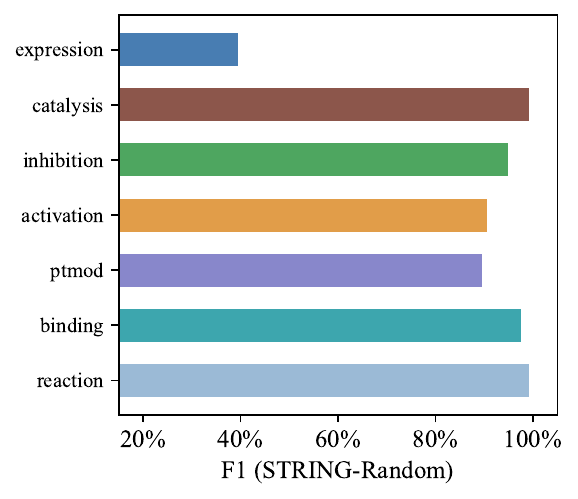}%
\label{fig:each_PPI_STRING-Random}}
\hfil
{\includegraphics[width=0.33\linewidth]{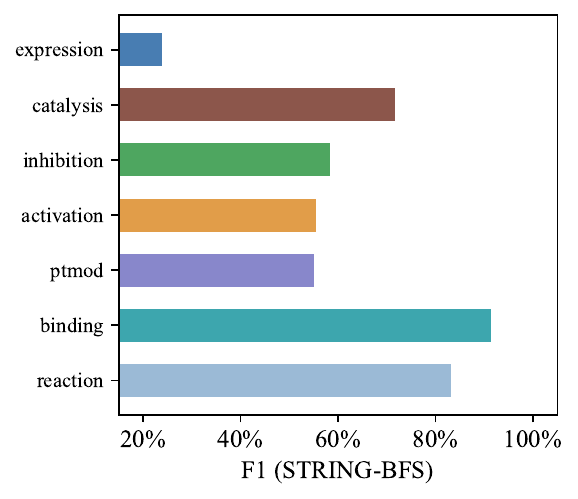}%
\label{fig:each_PPI_STRING-BFS}}
\hfil
{\includegraphics[width=0.33\linewidth]{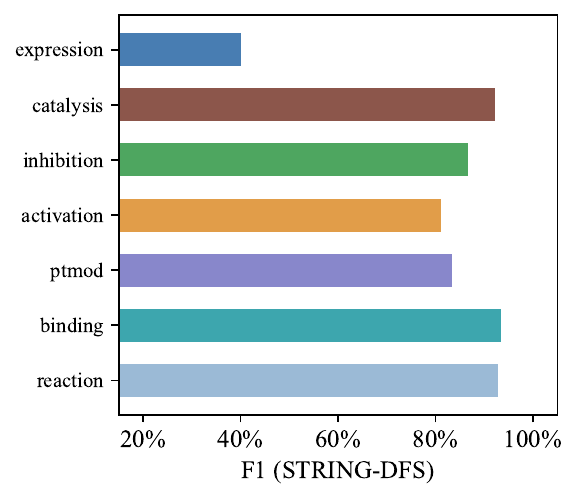}%
\label{fig:each_PPI_STRING-DFS}}
\caption{F1 scores for each PPI in benchmark datasets.}
\label{fig:each_PPI}
\end{figure*}

Based on the F1 scores from Table~\ref{tab:each_PPI}, we plot comparative histograms across benchmark datasets and partition schemes in Figure~\ref{fig:each_PPI}, to more directly convey the difficulty of identification for each type. For instance, it is clear that the categories of \textit{reaction}, \textit{binding}, and \textit{catalysis} consistently demonstrate robust classification performance across all experimental settings. This observation can be attributed to two factors: (1) These three categories constitute a higher proportion of the training data, allowing the model to extract corresponding interaction cues; (2) Their biological functions typically come with distinct molecular characteristics, making it easier for the model to capture discriminative patterns. In contrast, the \textit{expression} category has the lowest F1 scores across the three datasets, further supporting the hypothesis of insufficient learning for the minority class.

\subsection{Subset in-depth exploration}
\begin{figure*}[htbp]
\centering
{\includegraphics[width=0.33\linewidth]{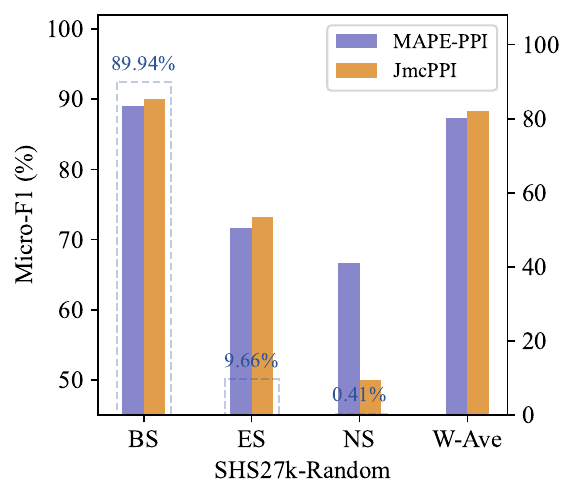}%
\label{fig:subset_comparison_SHS27k-Random}}
\hfil
{\includegraphics[width=0.33\linewidth]{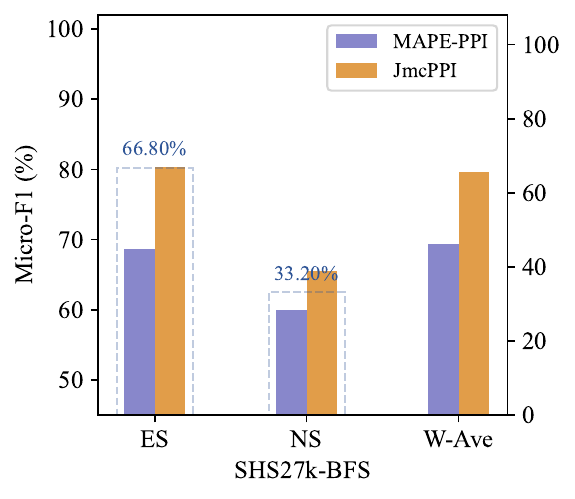}%
\label{fig:subset_comparison_SHS27k-BFS}}
\hfil
{\includegraphics[width=0.33\linewidth]{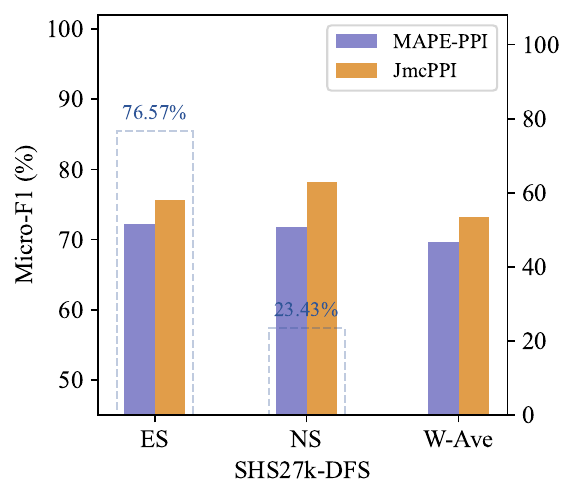}%
\label{fig:subset_comparison_SHS27k-DFS}}
\vfil
{\includegraphics[width=0.33\linewidth]{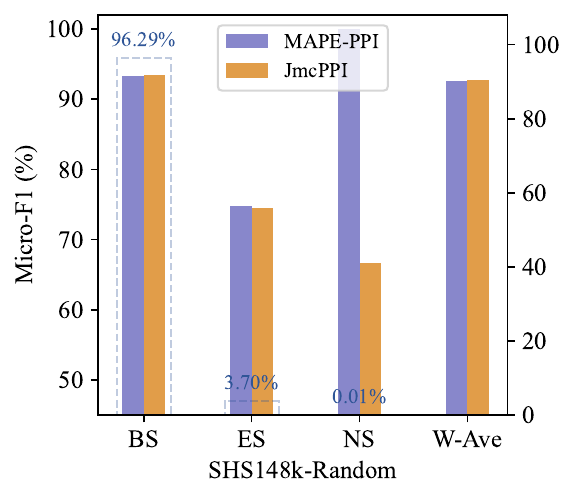}%
\label{fig:subset_comparison_SHS148k-Random}}
\hfil
{\includegraphics[width=0.33\linewidth]{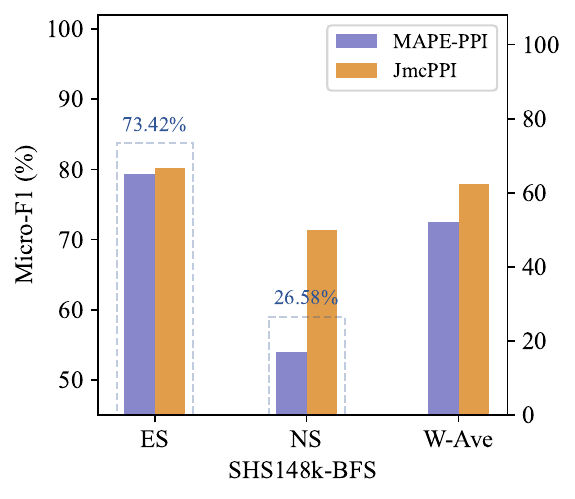}%
\label{fig:subset_comparison_SHS148k-BFS}}
\hfil
{\includegraphics[width=0.33\linewidth]{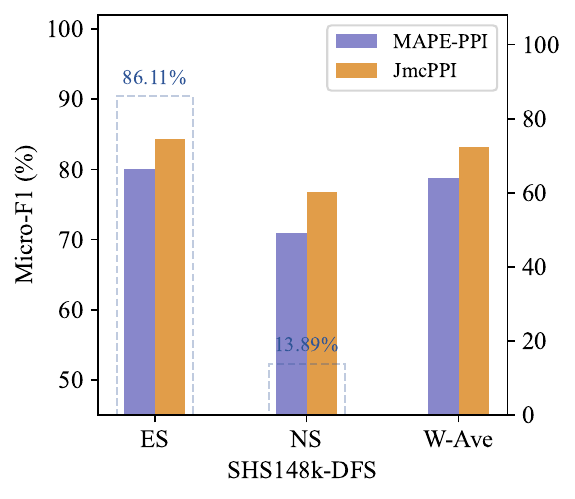}%
\label{fig:subset_comparison_SHS148k-DFS}}
\vfil
{\includegraphics[width=0.33\linewidth]{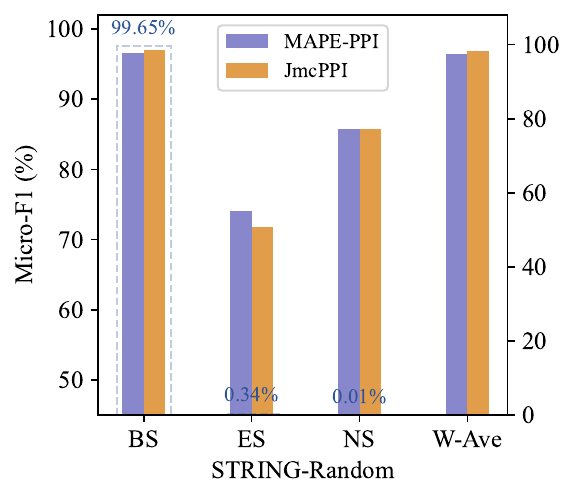}%
\label{fig:subset_comparison_STRING-Random}}
\hfil
{\includegraphics[width=0.33\linewidth]{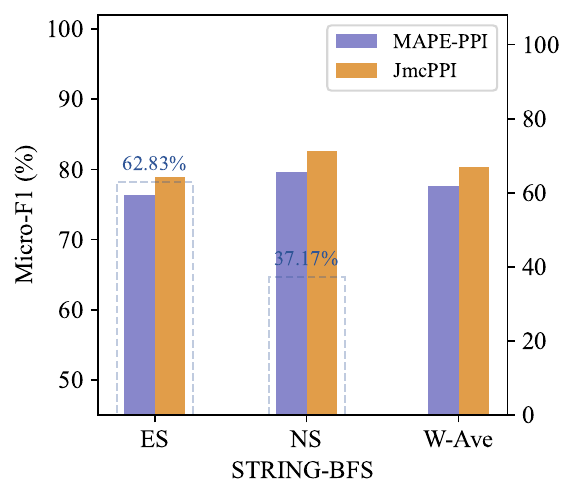}%
\label{fig:subset_comparison_STRING-BFS}}
\hfil
{\includegraphics[width=0.33\linewidth]{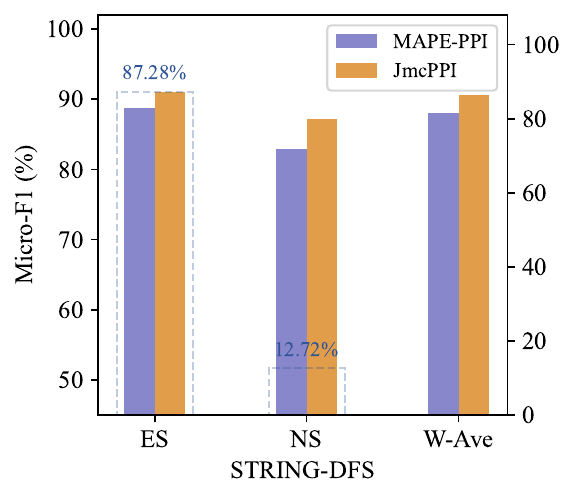}%
\label{fig:subset_comparison_STRING-DFS}}
\caption{Comparison of JmcPPI and MAPE-PPI on different subsets. Dashed boxe indicates the percentage of the corresponding subset in the whole dataset.}
\label{fig:subset_comparison}
\end{figure*}

To thoroughly investigate the generalization capability of JmcPPI, we partitioned the test set into three subsets: BS subset (both proteins in a pair exist in the training set), ES subset (one protein in a pair exists in the training set), and NS subset (neither protein in a pair exists in the labeled training set). Figure~\ref{fig:subset_comparison} illustrates the performance comparison between JmcPPI and MAPE-PPI across these subsets. For all benchmark datasets under BFS and DFS partitions, the test sets exclusively comprise ES and NS subsets. In contrast, the Random partition results in test sets dominated by the BS subset (accounting for 89.94\%, 96.29\%, and 99.65\% in SHS27k, SHS148k, and STRING, respectively), with minimal contributions from ES/NS subsets. Since models tend to leverage prior knowledge from seen proteins, this observation confirms that BFS and DFS partitions generate more challenging test scenarios compared to Random partition.

Experimental results reveal that JmcPPI significantly outperforms MAPE-PPI under BFS and DFS partitions, while achieving marginally higher weighted F1 scores under the Random partition. On the SHS27k dataset with BFS partition, JmcPPI attains micro-F1 scores of approximately 80\% (ES) and 78\% (NS), surpassing MAPE-PPI by 11\% and 6\%, respectively. Under the DFS partition, JmcPPI maintains stable advantages (ES: 76\% vs. 72\%; NS: 66\% vs. 60\%). This superiority is further validated in larger dataset. JmcPPI achieves weighted F1 scores of about 78\% (BFS) and 83\% (DFS) on SHS148k, which is an improvement of more than 5\%/4\% over MAPE-PPI. Similar trends are observed on the STRING dataset, where JmcPPI sustains micro-F1 values of around 83\% (BFS) and 87\% (DFS) for the NS subset. These findings demonstrate that our proposed hierarchical architecture, which decouples PPI prediction into residue structure encoding and protein interaction inference, effectively captures interaction patterns of unknown proteins, enabling reliable predictions in real-world open scenarios.

\subsection{Ablation study}
In this subsection, we validate the contributions of each module in the proposed approach through systematic ablation experiments, with experimental designs spanning three critical dimensions: (1) effectiveness investigation of reconstruction tasks, analyzing the differential impacts of standard reconstruction versus masked reconstruction on model performance; (2) effectiveness examination of contrastive learning mechanism, studying representation optimization effects between single-view contrastive learning and multi-graph contrastive learning; (3) sensitivity analysis of contrastive task types, investigating the functional mechanisms of edge perturbation contrast, node perturbation contrast, and non-perturbed contrast on prediction performance.

\begin{table*}[htbp]
\centering
\scriptsize
\setlength{\tabcolsep}{5.4pt}
\caption{\label{tab:wore}Micro-F1 score for removing residue reconstruction task. Here, w/o $\mathcal{L}_{\text{RE}}$ and w/o $\mathcal{L}_{\text{MSRE}}$ denote the ablation of the standard reconstruction and masked reconstruction tasks, respectively; w/o $\mathcal{L}_{\text{RE}}$\&$\mathcal{L}_{\text{MSRE}}$ indicates the simultaneous removal of both reconstruction tasks, i.e., complete removal of reconstruction tasks.}
\begin{tabular}{cccccccccccccc}
\toprule
\multirow{2}{*}{Method}  & \multicolumn{4}{c}{SHS27k} &\multicolumn{4}{c}{SHS148k} &\multicolumn{4}{c}{STRING} & \multirow{2}{*}{Average}\\
\cmidrule(lr){2-5}  \cmidrule(lr){6-9} \cmidrule(lr){10-13}
     & Random & BFS & DFS & Average & Random & BFS & DFS & Average & Random & BFS & DFS & Average & \\
\midrule
JmcPPI & {88.80} & {79.65} & {73.51} & {80.65} & {93.00} & {77.66} & {83.29} & {84.65} & {96.94} & {81.53} & {90.59} & {89.69} & {85.00} \\
\midrule
w/o $\mathcal{L}_{\text{RE}}$ &88.08 &78.31 &70.83 &79.07 &92.38 &77.03 &83.52 &84.31 &96.89 &81.23 &90.70 &89.61 &84.33 \\
w/o $\mathcal{L}_{\text{MSRE}}$ &87.87 &77.90 &71.98 &79.25 &92.84 &77.48 &82.97 &84.43 &96.91 &80.99 &90.69 &89.53 &84.40\\
w/o $\mathcal{L}_{\text{RE}}$\&$\mathcal{L}_{\text{MSRE}}$ &68.54 &67.19 &64.67 &66.80 &75.32 &66.26 &69.75 &70.44 &96.25 &80.87 &85.55 &87.56 &74.93\\
\bottomrule
\end{tabular}
\end{table*}
To validate the effectiveness of reconstruction tasks in the residue structure encoding stage, we design three experiments: removing the standard reconstruction task, removing the masked reconstruction task, and removing both reconstruction tasks simultaneously. The experimental results (Table~\ref{tab:wore}) show that the model's average performance declines to varying degrees when any reconstruction task is removed. Specifically, on the SHS27k dataset, removing the standard reconstruction task causes a 1.58\% decrease in the model's average micro-F1 score, while smaller performance drops are observed on SHS148k and STRING datasets (0.34\% and 0.08\%, respectively). Similarly, removing the masked reconstruction task results in micro-F1 score reductions of 1.40\%, 0.22\%, and 0.16\% on SHS27k, SHS148k, and STRING datasets. When both reconstruction tasks are removed, the model performance deteriorates significantly. For instance, under the Random, BFS, and DFS partition schemes on the SHS27k dataset, the performance decreases by 20.26\%, 12.46\%, and 8.84\%, respectively. For SHS148k and STRING datasets, the average micro-F1 scores across all partition schemes decline by 14.21\% and 2.13\%, respectively. These results indicate that the reconstruction tasks in the residue structure encoding stage enable the model to mine internal structural information of proteins, thereby enriching protein feature representations and critically contributing to performance enhancement.

\begin{figure*}[htbp]
\centering
{\includegraphics[width=0.33\linewidth]{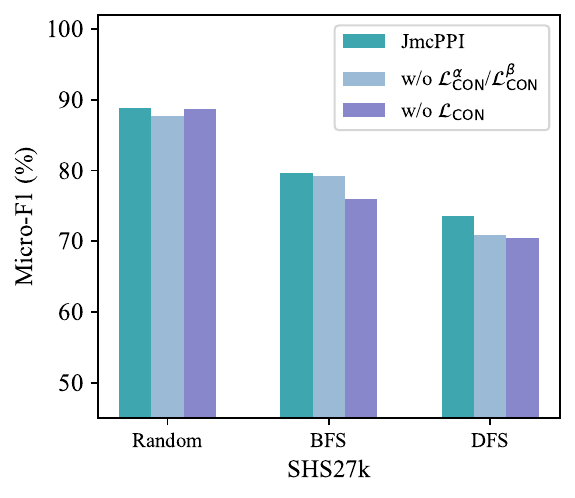}%
\label{fig:wocon_SHS27k}}
\hfil
{\includegraphics[width=0.33\linewidth]{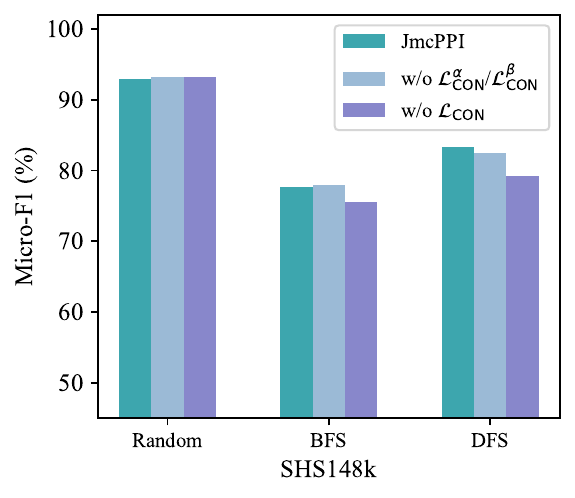}%
\label{fig:wocon_SHS148k}}
\hfil
{\includegraphics[width=0.33\linewidth]{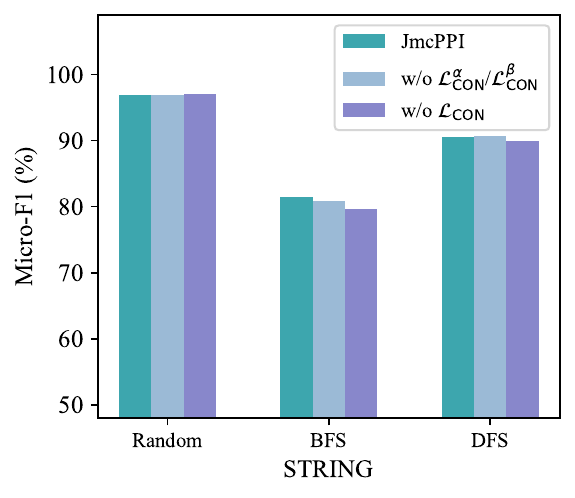}%
\label{fig:wocon_STRING}}
\caption{Performance changes for removal contrastive learning tasks. Here, w/o $\mathcal{L}_{\text{CON}}^\alpha$/$\mathcal{L}_{\text{CON}}^\beta$ refers to the ablation of single-view contrastive learning, and w/o $\mathcal{L}_{\text{CON}}$ signifies the removal of the entire multi-view contrastive learning.}
\label{fig:wocon}
\end{figure*}
We conduct two experiments to ascertain the effectiveness of contrastive learning tasks in the protein interaction inference phase. The first experiment involves the removal of single-view contrastive learning, and the second experiment includes the elimination of the entire multi-view contrastive learning. The results, as illustrated in Figure~\ref{fig:wocon}, indicate that the performance of the proposed model declines across most dataset configurations upon the removal of these contrastive learning tasks. On the SHS27k dataset, the removal of single-view contrastive learning leads to a decrease in model performance across all data partition schemes, with the most significant drop observed under the DFS partition; the elimination of multi-view contrastive learning results in notable performance declines under BFS and DFS data partitions. On the SHS148k dataset, the removal of single-view contrastive learning maintains comparable performance to the original model under Random and BFS partitions, but a decrease is observed under the DFS partition; the removal of multi-view contrastive learning causes severe performance degradations under both BFS and DFS partitions. On the STRING dataset, the removal of single-view contrastive learning results in a performance decline under the BFS partition, while performance remains similar to the original under the other two partition schemes; the elimination of multi-view contrastive learning leads to performance decreases under BFS and DFS partitions. Overall, the removal of the entire multi-view contrastive learning significantly impacts the model's performance under BFS and DFS partitions across all datasets. This suggests that under challenging data partition conditions, contrastive learning tasks in the protein interaction inference phase effectively enhance the model's ability to capture extrinsic interaction information for unknown proteins, thereby bolstering the model's overall performance.

\begin{figure*}[htbp]
\centering
{\includegraphics[width=0.33\linewidth]{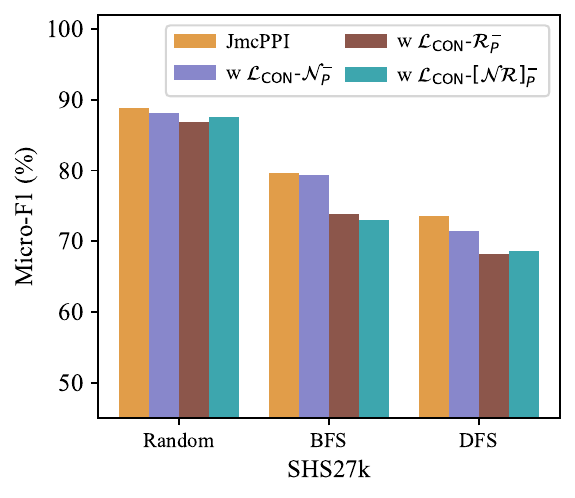}%
\label{fig:wcon-nrp_SHS27k}}
\hfil
{\includegraphics[width=0.33\linewidth]{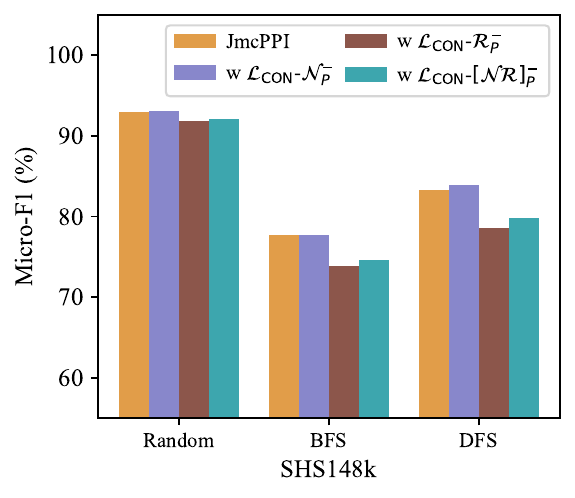}%
\label{fig:wcon-nrp_SHS148k}}
\hfil
{\includegraphics[width=0.33\linewidth]{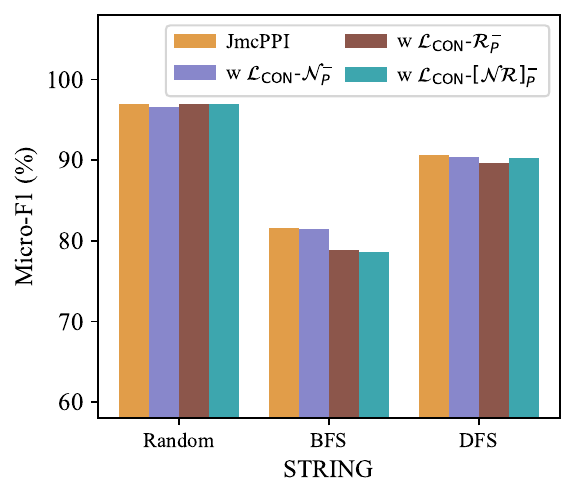}%
\label{fig:wcon-nrp_STRING}}
\caption{Performance comparison with different contrastive learning. Here, w $\mathcal{L}_{\text{CON}}$-$\mathcal{N}_P^{-}$ refers to multi-graph contrastive learning without node perturbation, w $\mathcal{L}_{\text{CON}}$-$\mathcal{R}_P^{-}$ refers to multi-graph contrastive learning without edge perturbation, and w $\mathcal{L}_{\text{CON}}$-$[\mathcal{NR}]_P^{-}$ refers to multi-graph contrastive learning without any perturbation.}
\label{fig:wcon-nrp}
\end{figure*}
To investigate the impact of different perturbation manners on the efficacy of contrastive learning, three sets of experiments are designed: (1) multi-graph contrastive learning based on non-perturbed nodes; (2) multi-graph contrastive learning based on non-perturbed edges; and (3) multi-graph contrastive learning based on non-perturbed nodes and edges. The experimental outcomes, as depicted in Figure~\ref{fig:wcon-nrp}, assess the performance of different strategies using the micro-F1 score. In the SHS27k dataset, contrastive learning with non-perturbed nodes leads to slight decreases in performance under Random and BFS partitions, while there is a significant drop in performance under the DFS partition; contrastive learning with non-perturbed edges results in noticeable declines in performance across all three partition schemes; when both nodes and edges remain non-perturbed, there is significant reductions in performance under BFS and DFS partitions. In the SHS148k dataset, contrastive learning with non-perturbed nodes maintains performance comparable to or slightly improved over the original model across all three partition schemes; contrastive learning with non-perturbed edges significantly decreases performance across all three partition schemes; and when both nodes and edges are non-perturbed, performance also significantly declines. In the STRING dataset, contrastive learning with non-perturbed nodes keeps performance essentially unchanged across all three partition schemes; contrastive learning with non-perturbed edges leads to drops in performance under BFS and DFS partitions; and when both nodes and edges remain non-perturbed, performance is comparable to the original model under Random and DFS partitions, while it decreases under the BFS partition. In summary, the multi-graph contrastive learning strategies with non-perturbed nodes or edges leads to performance reductions under certain partition schemes of benchmark datasets. This indicates that moderate node or edge perturbation is crucial for enhancing the effectiveness of contrastive learning, especially in challenging data partition scenarios.

\subsection{PPI embedding visualization}
\begin{figure*}[htbp]
\centering
{\includegraphics[width=0.32\linewidth]{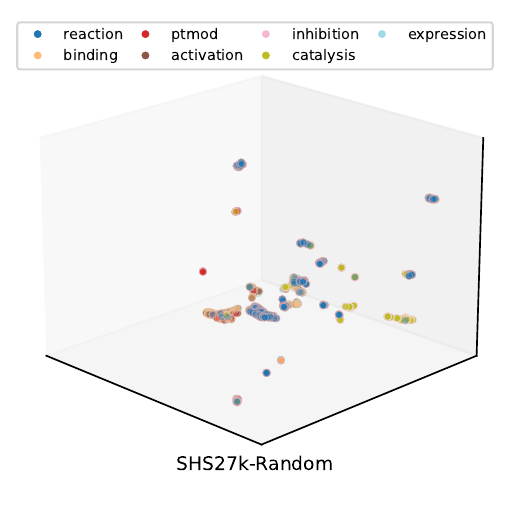}%
\label{fig:visual_SHS27k-Random}}
\hfil
{\includegraphics[width=0.32\linewidth]{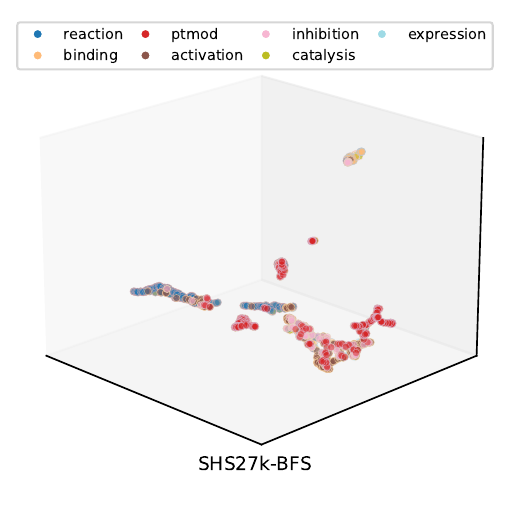}%
\label{fig:visual_SHS27k-BFS}}
\hfil
{\includegraphics[width=0.32\linewidth]{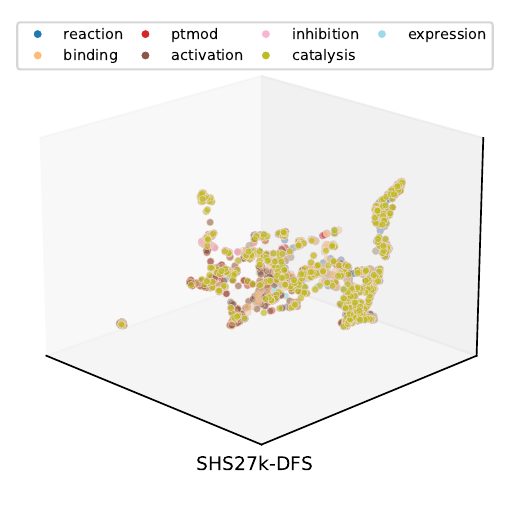}%
\label{fig:visual_SHS27k-DFS}}
\vfil
{\includegraphics[width=0.32\linewidth]{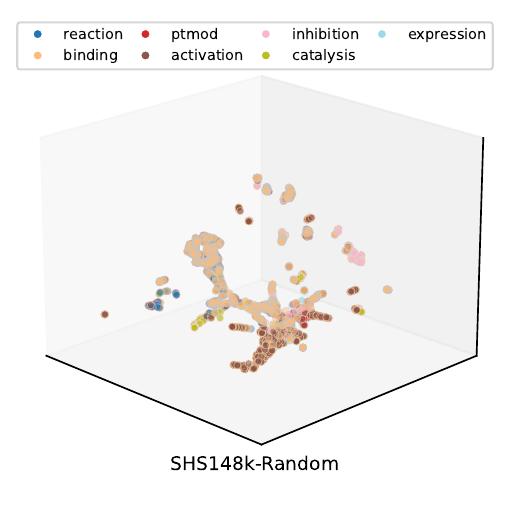}%
\label{fig:visual_SHS148k-Random}}
\hfil
{\includegraphics[width=0.32\linewidth]{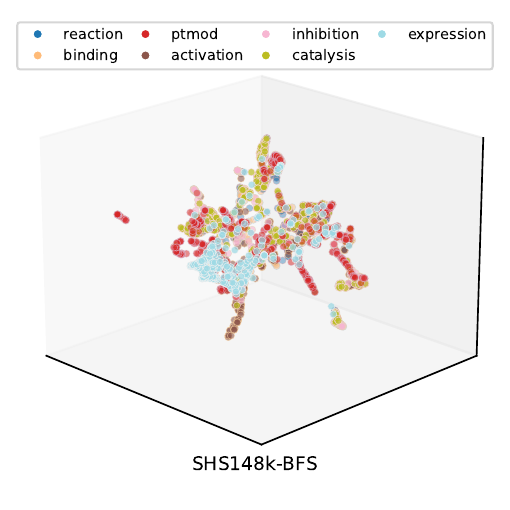}%
\label{fig:visual_SHS148k-BFS}}
\hfil
{\includegraphics[width=0.32\linewidth]{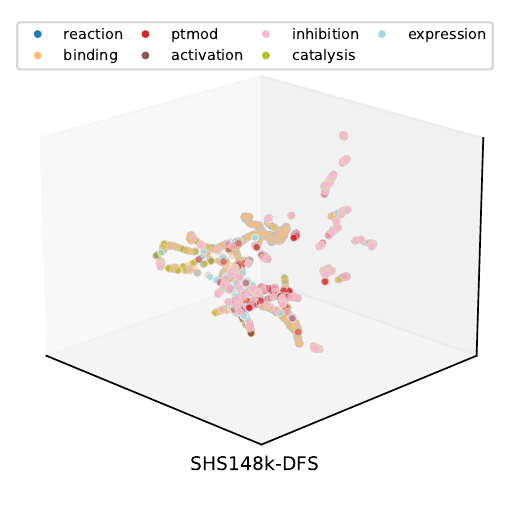}%
\label{fig:visual_SHS148k-DFS}}
\vfil
{\includegraphics[width=0.32\linewidth]{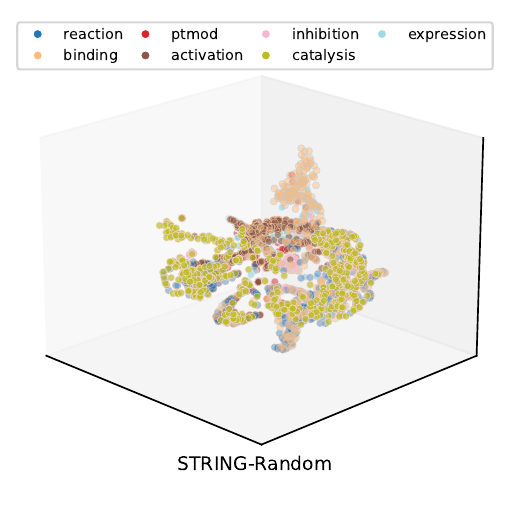}%
\label{fig:visual_STRING-Random}}
\hfil
{\includegraphics[width=0.32\linewidth]{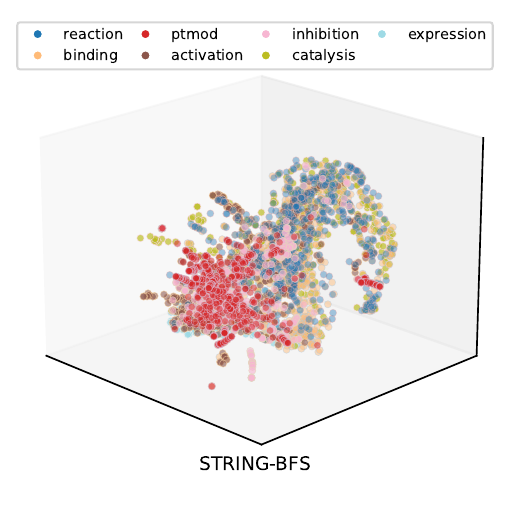}%
\label{fig:visual_STRING-BFS}}
\hfil
{\includegraphics[width=0.32\linewidth]{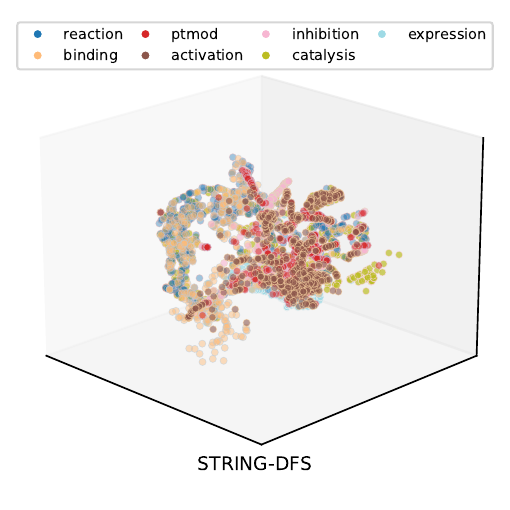}%
\label{fig:visual_STRING-DFS}}
\caption{Visualization of PPI embeddings in benchmark datasets.}
\label{fig:visual}
\end{figure*}

As shown in Figure~\ref{fig:visual}, we utilize the UMAP~\cite{mcinnes2018umap} tool to conduct a visual analysis of the PPI embeddings within benchmark datasets. Given that PPI prediction is a multi-label classification task, where each sample may correspond to multiple PPI type labels, we treat each PPI type as an independent subject during the visualization process, plotting points for samples corresponding to that type. Consequently, the same sample may appear multiple times in the figure as points of different colors, and these points may overlap. Considering that the sample size of the STRING dataset is significantly larger than those of the SHS27k and SHS148k datasets, to ensure clarity and interpretability of the visualization, we impose a maximum sample limit for each PPI type (no more than 500 samples per type) and perform random sampling within the dataset. The visualization results indicate that our proposed JmcPPI method can to some extent cluster PPI samples of the same type together, demonstrating the method's high accuracy in predicting protein interaction types.

\subsection{Impact of different protein dimensions}
\begin{figure*}[htbp]
\centering
{\includegraphics[width=0.33\linewidth]{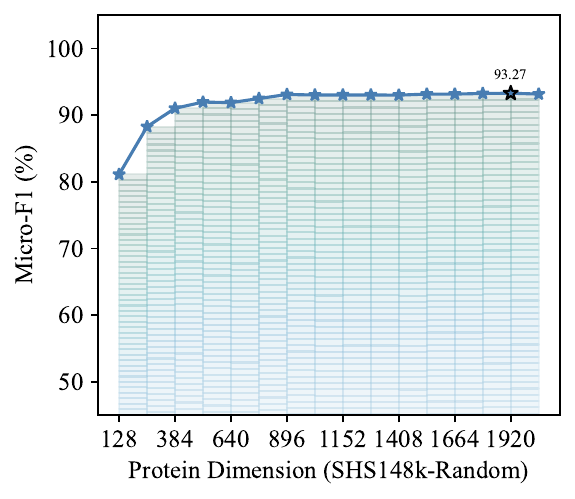}%
\label{fig:prot_hidden_dim_SHS148k-Random}}
\hfil
{\includegraphics[width=0.33\linewidth]{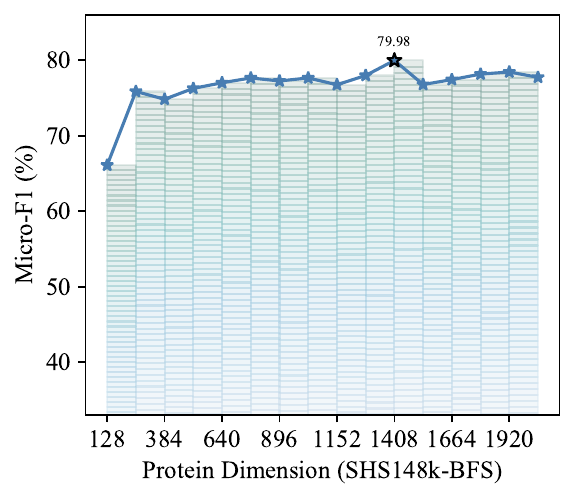}%
\label{fig:prot_hidden_dim_SHS148k-BFS}}
\hfil
{\includegraphics[width=0.33\linewidth]{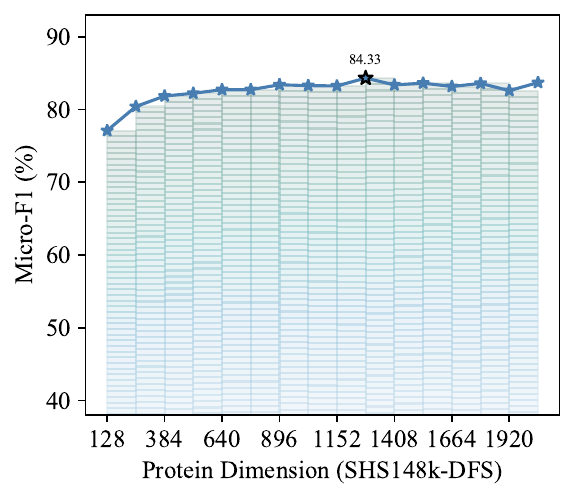}%
\label{fig:prot_hidden_dim_SHS148k-DFS}}
\caption{Performance discrepancies due to different protein dimensions on the SHS148k dataset.}
\label{fig:prot_hidden_dim_SHS148k}
\end{figure*}

Figure~\ref{fig:prot_hidden_dim_SHS148k} illustrates the impact of different protein dimensions on model performance on the SHS148k dataset. We set the protein dimensions to range from 128 to 2048, recording the micro-F1 score for each experiment at intervals of 128 dimensions. The experimental results indicate that when the protein dimension is set to 128, the model's performance is at its lowest across Random, BFS, and DFS partitions. Under the Random partition, the model's performance significantly improves with the increase in protein dimension, stabilizing when the dimension reaches 896, after which there are slight fluctuations, ultimately achieving the highest micro-F1 score of 93.27\% at a dimension of 1920. Under the BFS partition, the model's performance fluctuates upward with the increasing dimension, reaching its peak performance at a dimension of 1408 with a micro-F1 score of 79.98\%. Under the DFS partition, the model's performance also improves with the increasing dimension but begins to fluctuate after reaching 896 dimensions, ultimately achieving the maximum micro-F1 score of 84.33\% at 1280 dimensions. It is worth noting that, to maintain consistency and comparability in the experiments, we did not adopt the optimal dimension for each partition in this study but instead used the default setting of MAPE-PPI, which is 1024 dimensions, as the uniform protein dimension. Overall, the choice of protein dimension does have a certain impact on model performance, and appropriately increasing the dimension can enhance the model's performance.

\subsection{Impact of distinct perturbation rates}
\begin{figure*}[htbp]
\centering
{\includegraphics[width=0.33\linewidth]{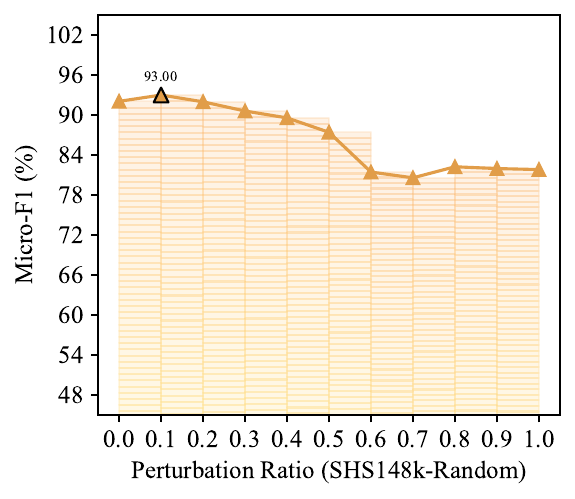}%
\label{fig:pert_ratio_SHS148k-Random}}
\hfil
{\includegraphics[width=0.33\linewidth]{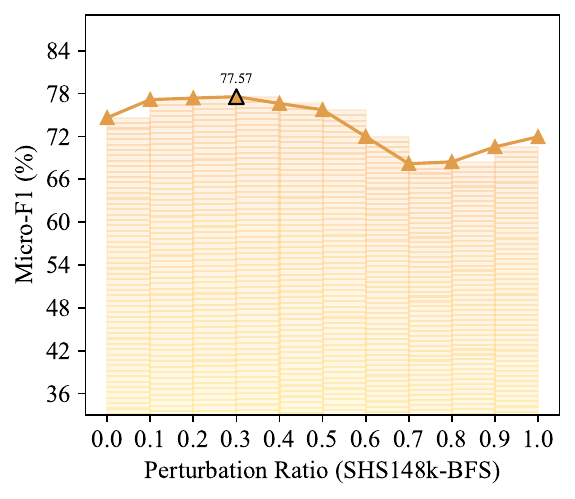}%
\label{fig:pert_ratio_SHS148k-BFS}}
\hfil
{\includegraphics[width=0.33\linewidth]{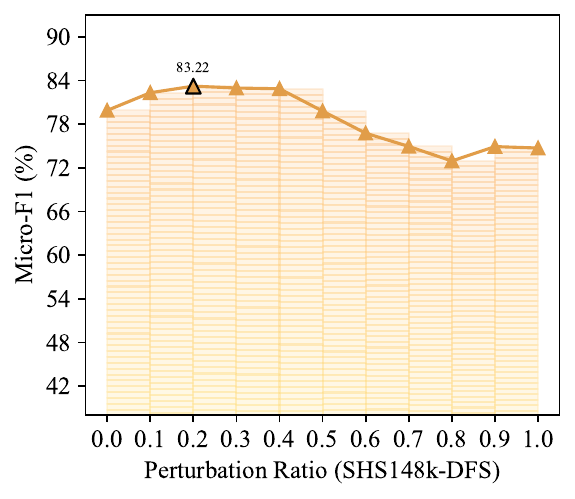}%
\label{fig:pert_ratio_SHS148k-DFS}}
\caption{Performance variations caused by distinct perturbation rates on the SHS148k dataset.}
\label{fig:pert_ratio_SHS148k}
\end{figure*}
To investigate the impact of varying perturbation rates on model performance, perturbation rates are set to range from 0 to 1 on the SHS148k dataset, with experiments conducted at intervals of 0.1, and the micro-F1 score is recorded for each experiment. As depicted in Figure~\ref{fig:pert_ratio_SHS148k}, the performance of the model under Random, BFS, and DFS partitions is significantly affected by the perturbation rate, exhibiting different trends under different partitions. For the Random partition, the model achieves the best performance at a perturbation rate of 0.1. As the perturbation rate increases beyond 0.1, the model performance shows a downward trend. Once the perturbation rate exceeds 0.7, the model performance gradually stabilizes. Under the BFS partition, the model performance improves with the increase of the perturbation rate, reaching its peak at a perturbation rate of 0.3. Subsequently, the model performance begins to decline, but there is a slight rebound as the perturbation rate continues to increase. Under the DFS partition, the model performance first enhances and then diminishes with the increase of the perturbation rate, attaining the optimal performance at a perturbation rate of 0.2. When the perturbation rate exceeds 0.8, the model performance begins to fluctuate. In general, the perturbation rate significantly influences model performance, and the model's sensitivity to the perturbation rate varies across different partitions. Additionally, a lower perturbation rate (e.g., between 0.1 and 0.3) is more conducive to enhancing model performance, whereas an excessively high perturbation rate may lead to performance decline or fluctuation.

\section{Conclusion}\label{sec:Conclusion}
To address the challenges faced by current PPI prediction methodologies, we propose an innovative approach driven by protein structure, known as JmcPPI. This method delineates the PPI prediction task into two distinct phases, i.e., residue structure encoding and protein interaction inference. In the initial phase, JmcPPI constructs a heterogeneous graph based on the topological structure of residues within proteins and integrates a heterogeneous graph attention mechanism with residue reconstruction task to efficiently extract internal structural information of proteins. In the subsequent phase, JmcPPI formulates the PPI network as a protein interaction graph and synergistically employs graph neural network and contrastive learning task to deeply mine the extrinsic interaction information of proteins. We have conducted extensive experiments across multiple benchmark datasets, encompassing model comparison, generalizability analysis, and ablation study. The experimental outcomes demonstrate that JmcPPI outperforms the state-of-the-art models under various data partitioning schemes. Looking ahead, we plan to apply JmcPPI to a broader spectrum of bioinformatics domains, such as protein-ligand affinity prediction and drug interaction prediction. Additionally, integrating amino acid sequences with biomedical text for interaction prediction presents a promising avenue for further exploration.

\section*{CRediT authorship contribution statement}
\textbf{Jiang Li}: Writing -- review \& editing, Writing -- original draft, Visualization, Validation, Software, Project administration, Methodology, Investigation, Formal analysis, Data curation, Conceptualization. \textbf{Xiaoping Wang}: Writing -- review \& editing, Supervision, Resources, Funding acquisition.

\section*{Declaration of competing interest}
The authors declare that they have no known competing financial interests or personal relationships that could have appeared to influence the work reported in this paper.

\section*{Data availability}
Data will be made available on request.

\section*{Acknowledgments}
This work was supported in part by the Interdisciplinary Research Program of HUST (2024JCYJ006), the National Natural Science Foundation of China (62236005 and 61936004), and the Fundamental Research Funds for the Central Universities of HUST (2024JYCXJJ067 and 2024JYCXJJ068). We thank the reviewers and editorial board for helpful comments that greatly improved the paper.

% \appendix
% \section{Example Appendix Section}


%%       \cite{<label>} ==> [1]

%%
% Example citation, See.




% \balance
% \bibliographystyle{elsarticle-num}
\bibliographystyle{elsarticle-harv}
\bibliography{jmcppi.bib}

\begin{thebibliography}{64}
\expandafter\ifx\csname natexlab\endcsname\relax\def\natexlab#1{#1}\fi
\providecommand{\url}[1]{\texttt{#1}}
\providecommand{\href}[2]{#2}
\providecommand{\path}[1]{#1}
\providecommand{\DOIprefix}{doi:}
\providecommand{\ArXivprefix}{arXiv:}
\providecommand{\URLprefix}{URL: }
\providecommand{\Pubmedprefix}{pmid:}
\providecommand{\doi}[1]{\href{http://dx.doi.org/#1}{\path{#1}}}
\providecommand{\Pubmed}[1]{\href{pmid:#1}{\path{#1}}}
\providecommand{\bibinfo}[2]{#2}
\ifx\xfnm\relax \def\xfnm[#1]{\unskip,\space#1}\fi
%Type = Article
\bibitem[{Asim et~al.(2022)Asim, Ibrahim, Malik, Dengel and
  Ahmed}]{asim2022adhppi}
\bibinfo{author}{Asim, M.N.}, \bibinfo{author}{Ibrahim, M.A.},
  \bibinfo{author}{Malik, M.I.}, \bibinfo{author}{Dengel, A.},
  \bibinfo{author}{Ahmed, S.}, \bibinfo{year}{2022}.
\newblock \bibinfo{title}{{ADH-PPI}: An attention-based deep hybrid model for
  protein-protein interaction prediction}.
\newblock \bibinfo{journal}{iScience} \bibinfo{volume}{25},
  \bibinfo{pages}{105169}.
\newblock \DOIprefix\doi{10.1016/j.isci.2022.105169}.
%Type = Article
\bibitem[{Bandyopadhyay and Mallick(2017)}]{7454750}
\bibinfo{author}{Bandyopadhyay, S.}, \bibinfo{author}{Mallick, K.},
  \bibinfo{year}{2017}.
\newblock \bibinfo{title}{A new feature vector based on gene ontology terms for
  protein-protein interaction prediction}.
\newblock \bibinfo{journal}{IEEE/ACM Transactions on Computational Biology and
  Bioinformatics} \bibinfo{volume}{14}, \bibinfo{pages}{762--770}.
\newblock \DOIprefix\doi{10.1109/TCBB.2016.2555304}.
%Type = Article
\bibitem[{Baranwal et~al.(2022)Baranwal, Magner, Saldinger, Turali-Emre,
  Elvati, Kozarekar, VanEpps, Kotov, Violi and Hero}]{Baranwal2022Struct2Graph}
\bibinfo{author}{Baranwal, M.}, \bibinfo{author}{Magner, A.},
  \bibinfo{author}{Saldinger, J.}, \bibinfo{author}{Turali-Emre, E.S.},
  \bibinfo{author}{Elvati, P.}, \bibinfo{author}{Kozarekar, S.},
  \bibinfo{author}{VanEpps, J.S.}, \bibinfo{author}{Kotov, N.A.},
  \bibinfo{author}{Violi, A.}, \bibinfo{author}{Hero, A.O.},
  \bibinfo{year}{2022}.
\newblock \bibinfo{title}{{Struct2Graph}: a graph attention network for
  structure based predictions of protein-protein interactions}.
\newblock \bibinfo{journal}{BMC Bioinformatics} \bibinfo{volume}{23},
  \bibinfo{pages}{370}.
\newblock \DOIprefix\doi{10.1186/s12859-022-04910-9}.
%Type = Article
\bibitem[{Chatterjee et~al.(2011)Chatterjee, Basu, Kundu, Nasipuri and
  Plewczynski}]{Chatterjee2011PPI_SVM}
\bibinfo{author}{Chatterjee, P.}, \bibinfo{author}{Basu, S.},
  \bibinfo{author}{Kundu, M.}, \bibinfo{author}{Nasipuri, M.},
  \bibinfo{author}{Plewczynski, D.}, \bibinfo{year}{2011}.
\newblock \bibinfo{title}{{PPI\_SVM}: prediction of protein-protein
  interactions using machine learning, domain-domain affinities and frequency
  tables}.
\newblock \bibinfo{journal}{Cellular and Molecular Biology Letters}
  \bibinfo{volume}{16}, \bibinfo{pages}{264--278}.
\newblock \DOIprefix\doi{10.2478/v10247-011-0053}.
%Type = Inproceedings
\bibitem[{Chen et~al.(2024a)Chen, Guo, Sun, Shao, Yuan, Lin and
  Zhang}]{ChenEVE2024}
\bibinfo{author}{Chen, J.}, \bibinfo{author}{Guo, L.}, \bibinfo{author}{Sun,
  J.}, \bibinfo{author}{Shao, S.}, \bibinfo{author}{Yuan, Z.},
  \bibinfo{author}{Lin, L.}, \bibinfo{author}{Zhang, D.},
  \bibinfo{year}{2024}a.
\newblock \bibinfo{title}{{EVE}: Efficient vision-language pre-training with
  masked prediction and modality-aware moe}, in:
  \bibinfo{booktitle}{Proceedings of the Thirty-Eighth AAAI Conference on
  Artificial Intelligence}, pp. \bibinfo{pages}{1110--1119}.
\newblock \DOIprefix\doi{10.1609/aaai.v38i2.27872}.
%Type = Inproceedings
\bibitem[{Chen et~al.(2023)Chen, Zhang, Li, Smola and
  Yang}]{chen-etal-2023-cheaper}
\bibinfo{author}{Chen, J.}, \bibinfo{author}{Zhang, A.}, \bibinfo{author}{Li,
  M.}, \bibinfo{author}{Smola, A.}, \bibinfo{author}{Yang, D.},
  \bibinfo{year}{2023}.
\newblock \bibinfo{title}{A cheaper and better diffusion language model with
  soft-masked noise}, in: \bibinfo{editor}{Bouamor, H.}, \bibinfo{editor}{Pino,
  J.}, \bibinfo{editor}{Bali, K.} (Eds.), \bibinfo{booktitle}{Proceedings of
  the 2023 Conference on Empirical Methods in Natural Language Processing},
  \bibinfo{publisher}{Association for Computational Linguistics},
  \bibinfo{address}{Singapore}. pp. \bibinfo{pages}{4765--4775}.
\newblock \DOIprefix\doi{10.18653/v1/2023.emnlp-main.289}.
%Type = Article
\bibitem[{Chen et~al.(2019)Chen, Ju, Zhou, Chen, Zhang, Chang, Zaniolo and
  Wang}]{10.1093/bioinformatics/btz328}
\bibinfo{author}{Chen, M.}, \bibinfo{author}{Ju, C.J.T.},
  \bibinfo{author}{Zhou, G.}, \bibinfo{author}{Chen, X.},
  \bibinfo{author}{Zhang, T.}, \bibinfo{author}{Chang, K.W.},
  \bibinfo{author}{Zaniolo, C.}, \bibinfo{author}{Wang, W.},
  \bibinfo{year}{2019}.
\newblock \bibinfo{title}{Multifaceted protein-protein interaction prediction
  based on siamese residual rcnn}.
\newblock \bibinfo{journal}{Bioinformatics} \bibinfo{volume}{35},
  \bibinfo{pages}{i305--i314}.
\newblock \DOIprefix\doi{10.1093/bioinformatics/btz328}.
%Type = Inproceedings
\bibitem[{Chen et~al.(2024b)Chen, Pertsemlidis and Chatterjee}]{chen2024pepmlm}
\bibinfo{author}{Chen, T.}, \bibinfo{author}{Pertsemlidis, S.},
  \bibinfo{author}{Chatterjee, P.}, \bibinfo{year}{2024}b.
\newblock \bibinfo{title}{Pep{MLM}: Target sequence-conditioned generation of
  peptide binders via masked language modeling}, in: \bibinfo{booktitle}{ICLR
  2024 Workshop on Generative and Experimental Perspectives for Biomolecular
  Design}, pp. \bibinfo{pages}{1--13}.
\newblock \URLprefix \url{https://openreview.net/forum?id=p6fz0rq7zu}.
%Type = Article
\bibitem[{Dhanuka et~al.(2023)Dhanuka, Singh and Tripathi}]{10049745}
\bibinfo{author}{Dhanuka, R.}, \bibinfo{author}{Singh, J.P.},
  \bibinfo{author}{Tripathi, A.}, \bibinfo{year}{2023}.
\newblock \bibinfo{title}{A comprehensive survey of deep learning techniques in
  protein function prediction}.
\newblock \bibinfo{journal}{IEEE/ACM Transactions on Computational Biology and
  Bioinformatics} \bibinfo{volume}{20}, \bibinfo{pages}{2291--2301}.
\newblock \DOIprefix\doi{10.1109/TCBB.2023.3247634}.
%Type = Inproceedings
\bibitem[{Dutta and Saha(2020)}]{dutta-saha-2020-amalgamation}
\bibinfo{author}{Dutta, P.}, \bibinfo{author}{Saha, S.}, \bibinfo{year}{2020}.
\newblock \bibinfo{title}{Amalgamation of protein sequence, structure and
  textual information for improving protein-protein interaction
  identification}, in: \bibinfo{editor}{Jurafsky, D.}, \bibinfo{editor}{Chai,
  J.}, \bibinfo{editor}{Schluter, N.}, \bibinfo{editor}{Tetreault, J.} (Eds.),
  \bibinfo{booktitle}{Proceedings of the 58th Annual Meeting of the Association
  for Computational Linguistics}, \bibinfo{publisher}{Association for
  Computational Linguistics}, \bibinfo{address}{Online}. pp.
  \bibinfo{pages}{6396--6407}.
\newblock \DOIprefix\doi{10.18653/v1/2020.acl-main.570}.
%Type = Article
\bibitem[{Fields and Sternglanz(1994)}]{Fields1994The}
\bibinfo{author}{Fields, S.}, \bibinfo{author}{Sternglanz, R.},
  \bibinfo{year}{1994}.
\newblock \bibinfo{title}{The two-hybrid system: an assay for protein-protein
  interactions}.
\newblock \bibinfo{journal}{Trends in Genetics} \bibinfo{volume}{10},
  \bibinfo{pages}{286--292}.
\newblock \DOIprefix\doi{10.1016/0168-9525(94)00076-6}.
%Type = Article
\bibitem[{Gao et~al.(2023)Gao, Jiang, Zhang, Jiang, Li, Zhao, Yang, Huang and
  Li}]{gao2023hierarchical}
\bibinfo{author}{Gao, Z.}, \bibinfo{author}{Jiang, C.}, \bibinfo{author}{Zhang,
  J.}, \bibinfo{author}{Jiang, X.}, \bibinfo{author}{Li, L.},
  \bibinfo{author}{Zhao, P.}, \bibinfo{author}{Yang, H.},
  \bibinfo{author}{Huang, Y.}, \bibinfo{author}{Li, J.}, \bibinfo{year}{2023}.
\newblock \bibinfo{title}{Hierarchical graph learning for protein-protein
  interaction}.
\newblock \bibinfo{journal}{Nature Communications} \bibinfo{volume}{14},
  \bibinfo{pages}{1093}.
%Type = Article
\bibitem[{Gavin et~al.(2002)Gavin, Bösche, Krause, Grandi, Marzioch, Bauer,
  Schultz, Rick, Michon, Cruciat, Remor, Höfert, Schelder, Brajenovic,
  Ruffner, Merino, Klein, Hudak, Dickson, Rudi, Gnau, Bauch, Bastuck, Huhse,
  Leutwein, Heurtier, Copley, Edelmann, Querfurth, Rybin, Drewes, Raida,
  Bouwmeester, Bork, Seraphin, Kuster, Neubauer and
  Superti-Furga}]{Gavin2002Functional}
\bibinfo{author}{Gavin, A.C.}, \bibinfo{author}{Bösche, M.},
  \bibinfo{author}{Krause, R.}, \bibinfo{author}{Grandi, P.},
  \bibinfo{author}{Marzioch, M.}, \bibinfo{author}{Bauer, A.},
  \bibinfo{author}{Schultz, J.}, \bibinfo{author}{Rick, J.M.},
  \bibinfo{author}{Michon, A.M.}, \bibinfo{author}{Cruciat, C.M.},
  \bibinfo{author}{Remor, M.}, \bibinfo{author}{Höfert, C.},
  \bibinfo{author}{Schelder, M.}, \bibinfo{author}{Brajenovic, M.},
  \bibinfo{author}{Ruffner, H.}, \bibinfo{author}{Merino, A.},
  \bibinfo{author}{Klein, K.}, \bibinfo{author}{Hudak, M.},
  \bibinfo{author}{Dickson, D.}, \bibinfo{author}{Rudi, T.},
  \bibinfo{author}{Gnau, V.}, \bibinfo{author}{Bauch, A.},
  \bibinfo{author}{Bastuck, S.}, \bibinfo{author}{Huhse, B.},
  \bibinfo{author}{Leutwein, C.}, \bibinfo{author}{Heurtier, M.A.},
  \bibinfo{author}{Copley, R.R.}, \bibinfo{author}{Edelmann, A.},
  \bibinfo{author}{Querfurth, E.}, \bibinfo{author}{Rybin, V.},
  \bibinfo{author}{Drewes, G.}, \bibinfo{author}{Raida, M.},
  \bibinfo{author}{Bouwmeester, T.}, \bibinfo{author}{Bork, P.},
  \bibinfo{author}{Seraphin, B.}, \bibinfo{author}{Kuster, B.},
  \bibinfo{author}{Neubauer, G.}, \bibinfo{author}{Superti-Furga, G.},
  \bibinfo{year}{2002}.
\newblock \bibinfo{title}{Functional organization of the yeast proteome by
  systematic analysis of protein complexes}.
\newblock \bibinfo{journal}{Nature} \bibinfo{volume}{415},
  \bibinfo{pages}{141--147}.
\newblock \DOIprefix\doi{10.1038/415141a}.
%Type = Article
\bibitem[{Guo et~al.(2008)Guo, Yu, Wen and Li}]{Guo2008Using}
\bibinfo{author}{Guo, Y.}, \bibinfo{author}{Yu, L.}, \bibinfo{author}{Wen, Z.},
  \bibinfo{author}{Li, M.}, \bibinfo{year}{2008}.
\newblock \bibinfo{title}{Using support vector machine combined with auto
  covariance to predict protein-protein interactions from protein sequences}.
\newblock \bibinfo{journal}{Nucleic Acids Research} \bibinfo{volume}{36},
  \bibinfo{pages}{3025--3030}.
\newblock \DOIprefix\doi{10.1093/nar/gkn159}.
%Type = Article
\bibitem[{Hakes et~al.(2008)Hakes, Pinney, Robertson and
  Lovell}]{Hakes2008Protein}
\bibinfo{author}{Hakes, L.}, \bibinfo{author}{Pinney, J.W.},
  \bibinfo{author}{Robertson, D.L.}, \bibinfo{author}{Lovell, S.C.},
  \bibinfo{year}{2008}.
\newblock \bibinfo{title}{Protein-protein interaction networks and
  biology--what's the connection?}
\newblock \bibinfo{journal}{Nature Biotechnology} \bibinfo{volume}{26},
  \bibinfo{pages}{69--72}.
\newblock \DOIprefix\doi{10.1038/nbt0108-69}.
%Type = Article
\bibitem[{Hashemifar et~al.(2018)Hashemifar, Neyshabur, Khan and
  Xu}]{10.1093/bioinformatics/bty573}
\bibinfo{author}{Hashemifar, S.}, \bibinfo{author}{Neyshabur, B.},
  \bibinfo{author}{Khan, A.A.}, \bibinfo{author}{Xu, J.}, \bibinfo{year}{2018}.
\newblock \bibinfo{title}{Predicting protein-protein interactions through
  sequence-based deep learning}.
\newblock \bibinfo{journal}{Bioinformatics} \bibinfo{volume}{34},
  \bibinfo{pages}{i802--i810}.
\newblock \DOIprefix\doi{10.1093/bioinformatics/bty573}.
%Type = Inproceedings
\bibitem[{Hendrycks et~al.(2019)Hendrycks, Mazeika, Kadavath and
  Song}]{NEURIPS2019_a2b15837}
\bibinfo{author}{Hendrycks, D.}, \bibinfo{author}{Mazeika, M.},
  \bibinfo{author}{Kadavath, S.}, \bibinfo{author}{Song, D.},
  \bibinfo{year}{2019}.
\newblock \bibinfo{title}{Using self-supervised learning can improve model
  robustness and uncertainty}, in: \bibinfo{editor}{Wallach, H.},
  \bibinfo{editor}{Larochelle, H.}, \bibinfo{editor}{Beygelzimer, A.},
  \bibinfo{editor}{d\textquotesingle Alch\'{e}-Buc, F.}, \bibinfo{editor}{Fox,
  E.}, \bibinfo{editor}{Garnett, R.} (Eds.), \bibinfo{booktitle}{Proceedings of
  the Thirty-Third Conference on Neural Information Processing Systems},
  \bibinfo{publisher}{Curran Associates, Inc.}. pp. \bibinfo{pages}{1--12}.
%Type = Article
\bibitem[{Ho et~al.(2002)Ho, Gruhler, Heilbut, Bader, Moore, Adams, Millar,
  Taylor, Bennett, Boutilier, Yang, Wolting, Donaldson, Schandorff, Shewnarane,
  Vo, Taggart, Goudreault, Muskat, Alfarano, Dewar, Lin, Michalickova, Willems,
  Sassi, Nielsen, Rasmussen, Andersen, Johansen, Hansen, Jespersen,
  Podtelejnikov, Nielsen, Crawford, Poulsen, Sørensen, Matthiesen,
  Hendrickson, Gleeson, Pawson, Moran, Durocher, Mann, Hogue, Figeys and
  Tyers}]{Ho2002Systematic}
\bibinfo{author}{Ho, Y.}, \bibinfo{author}{Gruhler, A.},
  \bibinfo{author}{Heilbut, A.}, \bibinfo{author}{Bader, G.D.},
  \bibinfo{author}{Moore, L.}, \bibinfo{author}{Adams, S.L.},
  \bibinfo{author}{Millar, A.}, \bibinfo{author}{Taylor, P.},
  \bibinfo{author}{Bennett, K.}, \bibinfo{author}{Boutilier, K.},
  \bibinfo{author}{Yang, L.}, \bibinfo{author}{Wolting, C.},
  \bibinfo{author}{Donaldson, I.}, \bibinfo{author}{Schandorff, S.},
  \bibinfo{author}{Shewnarane, J.}, \bibinfo{author}{Vo, M.},
  \bibinfo{author}{Taggart, J.}, \bibinfo{author}{Goudreault, M.},
  \bibinfo{author}{Muskat, B.}, \bibinfo{author}{Alfarano, C.},
  \bibinfo{author}{Dewar, D.}, \bibinfo{author}{Lin, Z.},
  \bibinfo{author}{Michalickova, K.}, \bibinfo{author}{Willems, A.R.},
  \bibinfo{author}{Sassi, H.}, \bibinfo{author}{Nielsen, P.A.},
  \bibinfo{author}{Rasmussen, K.J.}, \bibinfo{author}{Andersen, J.R.},
  \bibinfo{author}{Johansen, L.E.}, \bibinfo{author}{Hansen, L.H.},
  \bibinfo{author}{Jespersen, H.}, \bibinfo{author}{Podtelejnikov, A.},
  \bibinfo{author}{Nielsen, E.}, \bibinfo{author}{Crawford, J.},
  \bibinfo{author}{Poulsen, V.}, \bibinfo{author}{Sørensen, B.D.},
  \bibinfo{author}{Matthiesen, J.}, \bibinfo{author}{Hendrickson, R.C.},
  \bibinfo{author}{Gleeson, F.}, \bibinfo{author}{Pawson, T.},
  \bibinfo{author}{Moran, M.F.}, \bibinfo{author}{Durocher, D.},
  \bibinfo{author}{Mann, M.}, \bibinfo{author}{Hogue, C.W.V.},
  \bibinfo{author}{Figeys, D.}, \bibinfo{author}{Tyers, M.},
  \bibinfo{year}{2002}.
\newblock \bibinfo{title}{Systematic identification of protein complexes in
  saccharomyces cerevisiae by mass spectrometry}.
\newblock \bibinfo{journal}{Nature} \bibinfo{volume}{415},
  \bibinfo{pages}{180--183}.
\newblock \DOIprefix\doi{10.1038/415180a}.
%Type = Article
\bibitem[{Hou et~al.(2017)Hou, De~Geest, Vranken, Heringa and
  Feenstra}]{10.1093/bioinformatics/btx005}
\bibinfo{author}{Hou, Q.}, \bibinfo{author}{De~Geest, P.F.G.},
  \bibinfo{author}{Vranken, W.F.}, \bibinfo{author}{Heringa, J.},
  \bibinfo{author}{Feenstra, K.A.}, \bibinfo{year}{2017}.
\newblock \bibinfo{title}{Seeing the trees through the forest: sequence-based
  homo- and heteromeric protein-protein interaction sites prediction using
  random forest}.
\newblock \bibinfo{journal}{Bioinformatics} \bibinfo{volume}{33},
  \bibinfo{pages}{1479--1487}.
\newblock \DOIprefix\doi{10.1093/bioinformatics/btx005}.
%Type = Article
\bibitem[{Hu et~al.(2021)Hu, Feng, Zhou, Harrison and
  Chen}]{10.1093/bioinformatics/btab737}
\bibinfo{author}{Hu, X.}, \bibinfo{author}{Feng, C.}, \bibinfo{author}{Zhou,
  Y.}, \bibinfo{author}{Harrison, A.}, \bibinfo{author}{Chen, M.},
  \bibinfo{year}{2021}.
\newblock \bibinfo{title}{{DeepTrio}: a ternary prediction system for
  protein–protein interaction using mask multiple parallel convolutional
  neural networks}.
\newblock \bibinfo{journal}{Bioinformatics} \bibinfo{volume}{38},
  \bibinfo{pages}{694--702}.
\newblock \DOIprefix\doi{10.1093/bioinformatics/btab737}.
%Type = Article
\bibitem[{Huang et~al.(2020)Huang, Xiao, Glass, Zitnik and
  Sun}]{Huang2020SkipGNN}
\bibinfo{author}{Huang, K.}, \bibinfo{author}{Xiao, C.},
  \bibinfo{author}{Glass, L.M.}, \bibinfo{author}{Zitnik, M.},
  \bibinfo{author}{Sun, J.}, \bibinfo{year}{2020}.
\newblock \bibinfo{title}{{SkipGNN}: predicting molecular interactions with
  skip-graph networks}.
\newblock \bibinfo{journal}{Scientific Reports} \bibinfo{volume}{10},
  \bibinfo{pages}{21092}.
\newblock \DOIprefix\doi{10.1038/s41598-020-77766-9}.
%Type = Article
\bibitem[{Huang et~al.(2023)Huang, Wuchty, Zhou and
  Zhang}]{10.1093/bib/bbad020}
\bibinfo{author}{Huang, Y.}, \bibinfo{author}{Wuchty, S.},
  \bibinfo{author}{Zhou, Y.}, \bibinfo{author}{Zhang, Z.},
  \bibinfo{year}{2023}.
\newblock \bibinfo{title}{{SGPPI}: structure-aware prediction of
  protein-protein interactions in rigorous conditions with graph convolutional
  network}.
\newblock \bibinfo{journal}{Briefings in Bioinformatics} \bibinfo{volume}{24},
  \bibinfo{pages}{bbad020}.
\newblock \DOIprefix\doi{10.1093/bib/bbad020}.
%Type = Article
\bibitem[{Jha et~al.(2022)Jha, Saha and Singh}]{Jha2022Prediction}
\bibinfo{author}{Jha, K.}, \bibinfo{author}{Saha, S.}, \bibinfo{author}{Singh,
  H.}, \bibinfo{year}{2022}.
\newblock \bibinfo{title}{Prediction of protein--protein interaction using
  graph neural networks}.
\newblock \bibinfo{journal}{Scientific Reports} \bibinfo{volume}{12},
  \bibinfo{pages}{8360}.
\newblock \URLprefix \url{https://doi.org/10.1038/s41598-022-12201-9},
  \DOIprefix\doi{10.1038/s41598-022-12201-9}.
%Type = Article
\bibitem[{Jumper et~al.(2021)Jumper, Evans, Pritzel, Green, Figurnov,
  Ronneberger, Tunyasuvunakool, Bates, Žídek, Potapenko, Bridgland, Meyer,
  Kohl, Ballard, Cowie, Romera-Paredes, Nikolov, Jain, Adler, Back, Petersen,
  Reiman, Clancy, Zielinski, Steinegger, Pacholska, Berghammer, Bodenstein,
  Silver, Vinyals, Senior, Kavukcuoglu, Kohli and Hassabis}]{Jumper2021Highly}
\bibinfo{author}{Jumper, J.}, \bibinfo{author}{Evans, R.},
  \bibinfo{author}{Pritzel, A.}, \bibinfo{author}{Green, T.},
  \bibinfo{author}{Figurnov, M.}, \bibinfo{author}{Ronneberger, O.},
  \bibinfo{author}{Tunyasuvunakool, K.}, \bibinfo{author}{Bates, R.},
  \bibinfo{author}{Žídek, A.}, \bibinfo{author}{Potapenko, A.},
  \bibinfo{author}{Bridgland, A.}, \bibinfo{author}{Meyer, C.},
  \bibinfo{author}{Kohl, S.A.A.}, \bibinfo{author}{Ballard, A.J.},
  \bibinfo{author}{Cowie, A.}, \bibinfo{author}{Romera-Paredes, B.},
  \bibinfo{author}{Nikolov, S.}, \bibinfo{author}{Jain, R.},
  \bibinfo{author}{Adler, J.}, \bibinfo{author}{Back, T.},
  \bibinfo{author}{Petersen, S.}, \bibinfo{author}{Reiman, D.},
  \bibinfo{author}{Clancy, E.}, \bibinfo{author}{Zielinski, M.},
  \bibinfo{author}{Steinegger, M.}, \bibinfo{author}{Pacholska, M.},
  \bibinfo{author}{Berghammer, T.}, \bibinfo{author}{Bodenstein, S.},
  \bibinfo{author}{Silver, D.}, \bibinfo{author}{Vinyals, O.},
  \bibinfo{author}{Senior, A.W.}, \bibinfo{author}{Kavukcuoglu, K.},
  \bibinfo{author}{Kohli, P.}, \bibinfo{author}{Hassabis, D.},
  \bibinfo{year}{2021}.
\newblock \bibinfo{title}{Highly accurate protein structure prediction with
  alphafold}.
\newblock \bibinfo{journal}{Nature} \bibinfo{volume}{596},
  \bibinfo{pages}{583--589}.
\newblock \DOIprefix\doi{10.1038/s41586-021-03819-2}.
%Type = Article
\bibitem[{Kang et~al.(2021)Kang, Zhang, Liu, Huang and
  Yin}]{10.1093/bib/bbab513}
\bibinfo{author}{Kang, C.}, \bibinfo{author}{Zhang, H.}, \bibinfo{author}{Liu,
  Z.}, \bibinfo{author}{Huang, S.}, \bibinfo{author}{Yin, Y.},
  \bibinfo{year}{2021}.
\newblock \bibinfo{title}{{LR-GNN}: a graph neural network based on link
  representation for predicting molecular associations}.
\newblock \bibinfo{journal}{Briefings in Bioinformatics} \bibinfo{volume}{23},
  \bibinfo{pages}{bbab513}.
\newblock \DOIprefix\doi{10.1093/bib/bbab513}.
%Type = Article
\bibitem[{Kang et~al.(2023)Kang, Elofsson, Jiang, Huang, Yu and
  Li}]{10.1093/bioinformatics/btad052}
\bibinfo{author}{Kang, Y.}, \bibinfo{author}{Elofsson, A.},
  \bibinfo{author}{Jiang, Y.}, \bibinfo{author}{Huang, W.},
  \bibinfo{author}{Yu, M.}, \bibinfo{author}{Li, Z.}, \bibinfo{year}{2023}.
\newblock \bibinfo{title}{{AFTGAN}: prediction of multi-type ppi based on
  attention free transformer and graph attention network}.
\newblock \bibinfo{journal}{Bioinformatics} \bibinfo{volume}{39},
  \bibinfo{pages}{btad052}.
\newblock \DOIprefix\doi{10.1093/bioinformatics/btad052}.
%Type = Article
\bibitem[{Keskin et~al.(2008)Keskin, Tuncbag and
  Gursoy}]{Keskin2008Characterization}
\bibinfo{author}{Keskin, O.}, \bibinfo{author}{Tuncbag, N.},
  \bibinfo{author}{Gursoy, A.}, \bibinfo{year}{2008}.
\newblock \bibinfo{title}{Characterization and prediction of protein interfaces
  to infer protein-protein interaction networks}.
\newblock \bibinfo{journal}{Current Pharmaceutical Biotechnology}
  \bibinfo{volume}{9}, \bibinfo{pages}{67--76}.
\newblock \DOIprefix\doi{10.2174/138920108783955191}.
%Type = Article
\bibitem[{Kibar and Vingron(2023)}]{10.1002/prot.26486}
\bibinfo{author}{Kibar, G.}, \bibinfo{author}{Vingron, M.},
  \bibinfo{year}{2023}.
\newblock \bibinfo{title}{Prediction of protein-protein interactions using
  sequences of intrinsically disordered regions}.
\newblock \bibinfo{journal}{Proteins: Structure, Function, and Bioinformatics}
  \bibinfo{volume}{91}, \bibinfo{pages}{980--990}.
\newblock \DOIprefix\doi{10.1002/prot.26486}.
%Type = Inproceedings
\bibitem[{Kingma and Ba(2015)}]{KingmaB14Adam}
\bibinfo{author}{Kingma, D.P.}, \bibinfo{author}{Ba, J.}, \bibinfo{year}{2015}.
\newblock \bibinfo{title}{Adam: {A} method for stochastic optimization}, in:
  \bibinfo{editor}{Bengio, Y.}, \bibinfo{editor}{LeCun, Y.} (Eds.),
  \bibinfo{booktitle}{Proceedings of the 3rd International Conference on
  Learning Representations}, pp. \bibinfo{pages}{1--15}.
%Type = Inproceedings
\bibitem[{Kipf and Welling(2017)}]{kipf2017semisupervised}
\bibinfo{author}{Kipf, T.N.}, \bibinfo{author}{Welling, M.},
  \bibinfo{year}{2017}.
\newblock \bibinfo{title}{Semi-supervised classification with graph
  convolutional networks}, in: \bibinfo{booktitle}{Proceedings of the 5th
  International Conference on Learning Representations}, pp.
  \bibinfo{pages}{1--14}.
%Type = Article
\bibitem[{Li et~al.(2018)Li, Gong, Yu and Zhou}]{molecules23081923}
\bibinfo{author}{Li, H.}, \bibinfo{author}{Gong, X.J.}, \bibinfo{author}{Yu,
  H.}, \bibinfo{author}{Zhou, C.}, \bibinfo{year}{2018}.
\newblock \bibinfo{title}{Deep neural network based predictions of protein
  interactions using primary sequences}.
\newblock \bibinfo{journal}{Molecules} \bibinfo{volume}{23}.
\newblock \DOIprefix\doi{10.3390/molecules23081923}.
%Type = Article
\bibitem[{Li et~al.(2024a)Li, Wang, Lv and Zeng}]{10081075}
\bibinfo{author}{Li, J.}, \bibinfo{author}{Wang, X.}, \bibinfo{author}{Lv, G.},
  \bibinfo{author}{Zeng, Z.}, \bibinfo{year}{2024}a.
\newblock \bibinfo{title}{{GA2MIF}: Graph and attention based two-stage
  multi-source information fusion for conversational emotion detection}.
\newblock \bibinfo{journal}{IEEE Transactions on Affective Computing}
  \bibinfo{volume}{15}, \bibinfo{pages}{130--143}.
\newblock \DOIprefix\doi{10.1109/TAFFC.2023.3261279}.
%Type = Article
\bibitem[{Li et~al.(2024b)Li, Wang and Zeng}]{li2024tracing}
\bibinfo{author}{Li, J.}, \bibinfo{author}{Wang, X.}, \bibinfo{author}{Zeng,
  Z.}, \bibinfo{year}{2024}b.
\newblock \bibinfo{title}{Tracing intricate cues in dialogue: Joint graph
  structure and sentiment dynamics for multimodal emotion recognition}.
\newblock \bibinfo{journal}{arXiv preprint arXiv:2407.21536} .
%Type = Article
\bibitem[{Li et~al.(2007)Li, Lin, Wang and Liu}]{10.1093/bioinformatics/btl660}
\bibinfo{author}{Li, M.H.}, \bibinfo{author}{Lin, L.}, \bibinfo{author}{Wang,
  X.L.}, \bibinfo{author}{Liu, T.}, \bibinfo{year}{2007}.
\newblock \bibinfo{title}{Protein-protein interaction site prediction based on
  conditional random fields}.
\newblock \bibinfo{journal}{Bioinformatics} \bibinfo{volume}{23},
  \bibinfo{pages}{597--604}.
\newblock \DOIprefix\doi{10.1093/bioinformatics/btl660}.
%Type = Article
\bibitem[{Li et~al.(2022)Li, Han, Wang, Chen, Wang and Song}]{Li2022SDNN-PPI}
\bibinfo{author}{Li, X.}, \bibinfo{author}{Han, P.}, \bibinfo{author}{Wang,
  G.}, \bibinfo{author}{Chen, W.}, \bibinfo{author}{Wang, S.},
  \bibinfo{author}{Song, T.}, \bibinfo{year}{2022}.
\newblock \bibinfo{title}{{SDNN-PPI}: self-attention with deep neural network
  effect on protein-protein interaction prediction}.
\newblock \bibinfo{journal}{BMC Genomics} \bibinfo{volume}{23},
  \bibinfo{pages}{474}.
\newblock \DOIprefix\doi{10.1186/s12864-022-08687-2}.
%Type = Article
\bibitem[{Lin and Chen(2013)}]{10.1002/pmic.201200326}
\bibinfo{author}{Lin, X.}, \bibinfo{author}{Chen, X.w.}, \bibinfo{year}{2013}.
\newblock \bibinfo{title}{Heterogeneous data integration by tree-augmented
  naïve bayes for protein-protein interactions prediction}.
\newblock \bibinfo{journal}{PROTEOMICS} \bibinfo{volume}{13},
  \bibinfo{pages}{261--268}.
\newblock \DOIprefix\doi{10.1002/pmic.201200326}.
%Type = Article
\bibitem[{Liu et~al.(2023a)Liu, Jin, Pan, Zhou, Zheng, Xia and Yu}]{9770382}
\bibinfo{author}{Liu, Y.}, \bibinfo{author}{Jin, M.}, \bibinfo{author}{Pan,
  S.}, \bibinfo{author}{Zhou, C.}, \bibinfo{author}{Zheng, Y.},
  \bibinfo{author}{Xia, F.}, \bibinfo{author}{Yu, P.S.}, \bibinfo{year}{2023}a.
\newblock \bibinfo{title}{Graph self-supervised learning: A survey}.
\newblock \bibinfo{journal}{IEEE Transactions on Knowledge and Data
  Engineering} \bibinfo{volume}{35}, \bibinfo{pages}{5879--5900}.
\newblock \DOIprefix\doi{10.1109/TKDE.2022.3172903}.
%Type = Inproceedings
\bibitem[{Liu et~al.(2023b)Liu, Shi, Zhang, Zhang, Kawaguchi, Wang and
  Chua}]{liu2023rethinking}
\bibinfo{author}{Liu, Z.}, \bibinfo{author}{Shi, Y.}, \bibinfo{author}{Zhang,
  A.}, \bibinfo{author}{Zhang, E.}, \bibinfo{author}{Kawaguchi, K.},
  \bibinfo{author}{Wang, X.}, \bibinfo{author}{Chua, T.S.},
  \bibinfo{year}{2023}b.
\newblock \bibinfo{title}{Rethinking tokenizer and decoder in masked graph
  modeling for molecules}, in: \bibinfo{booktitle}{Proceedings of the
  Thirty-Seventh Conference on Neural Information Processing Systems}, pp.
  \bibinfo{pages}{1--22}.
\newblock \URLprefix \url{https://openreview.net/forum?id=fWLf8DV0fI}.
%Type = Inproceedings
\bibitem[{Lv et~al.(2021)Lv, Hu, Bi and Zhang}]{ijcai2021p506}
\bibinfo{author}{Lv, G.}, \bibinfo{author}{Hu, Z.}, \bibinfo{author}{Bi, Y.},
  \bibinfo{author}{Zhang, S.}, \bibinfo{year}{2021}.
\newblock \bibinfo{title}{Learning unknown from correlations: Graph neural
  network for inter-novel-protein interaction prediction}, in:
  \bibinfo{editor}{Zhou, Z.H.} (Ed.), \bibinfo{booktitle}{Proceedings of the
  Thirtieth International Joint Conference on Artificial Intelligence},
  \bibinfo{publisher}{International Joint Conferences on Artificial
  Intelligence Organization}. pp. \bibinfo{pages}{3677--3683}.
\newblock \DOIprefix\doi{10.24963/ijcai.2021/506}. \bibinfo{note}{main Track}.
%Type = Article
\bibitem[{McInnes et~al.(2018)McInnes, Healy, Saul and
  Gro{\ss}berger}]{mcinnes2018umap}
\bibinfo{author}{McInnes, L.}, \bibinfo{author}{Healy, J.},
  \bibinfo{author}{Saul, N.}, \bibinfo{author}{Gro{\ss}berger, L.},
  \bibinfo{year}{2018}.
\newblock \bibinfo{title}{Umap: Uniform manifold approximation and projection}.
\newblock \bibinfo{journal}{Journal of Open Source Software}
  \bibinfo{volume}{3}, \bibinfo{pages}{861}.
%Type = Article
\bibitem[{Murakami and Mizuguchi(2010)}]{10.1093/bioinformatics/btq302}
\bibinfo{author}{Murakami, Y.}, \bibinfo{author}{Mizuguchi, K.},
  \bibinfo{year}{2010}.
\newblock \bibinfo{title}{Applying the naïve bayes classifier with kernel
  density estimation to the prediction of protein-protein interaction sites}.
\newblock \bibinfo{journal}{Bioinformatics} \bibinfo{volume}{26},
  \bibinfo{pages}{1841--1848}.
\newblock \DOIprefix\doi{10.1093/bioinformatics/btq302}.
%Type = Article
\bibitem[{Peng et~al.(2025)Peng, Yuan, Zhang, Li, Dai, Wang, Wang and
  Wu}]{10872817}
\bibinfo{author}{Peng, T.}, \bibinfo{author}{Yuan, H.}, \bibinfo{author}{Zhang,
  Y.}, \bibinfo{author}{Li, Y.}, \bibinfo{author}{Dai, P.},
  \bibinfo{author}{Wang, Q.}, \bibinfo{author}{Wang, S.}, \bibinfo{author}{Wu,
  W.}, \bibinfo{year}{2025}.
\newblock \bibinfo{title}{{TagRec}: Temporal-aware graph contrastive learning
  with theoretical augmentation for sequential recommendation}.
\newblock \bibinfo{journal}{IEEE Transactions on Knowledge and Data
  Engineering} ,
  \bibinfo{pages}{1--14}\DOIprefix\doi{10.1109/TKDE.2025.3538706}.
%Type = Article
\bibitem[{Rahmani et~al.(2023)Rahmani, Baghbani, Bouguila and
  Patterson}]{10077454}
\bibinfo{author}{Rahmani, S.}, \bibinfo{author}{Baghbani, A.},
  \bibinfo{author}{Bouguila, N.}, \bibinfo{author}{Patterson, Z.},
  \bibinfo{year}{2023}.
\newblock \bibinfo{title}{Graph neural networks for intelligent transportation
  systems: A survey}.
\newblock \bibinfo{journal}{IEEE Transactions on Intelligent Transportation
  Systems} \bibinfo{volume}{24}, \bibinfo{pages}{8846--8885}.
\newblock \DOIprefix\doi{10.1109/TITS.2023.3257759}.
%Type = Article
\bibitem[{Sun et~al.(2017)Sun, Zhou, Lai and Pei}]{Sun2017Sequence-based}
\bibinfo{author}{Sun, T.}, \bibinfo{author}{Zhou, B.}, \bibinfo{author}{Lai,
  L.}, \bibinfo{author}{Pei, J.}, \bibinfo{year}{2017}.
\newblock \bibinfo{title}{Sequence-based prediction of protein protein
  interaction using a deep-learning algorithm}.
\newblock \bibinfo{journal}{BMC Bioinformatics} \bibinfo{volume}{18},
  \bibinfo{pages}{277}.
\newblock \DOIprefix\doi{10.1186/s12859-017-1700-2}.
%Type = Article
\bibitem[{Szklarczyk et~al.(2019)Szklarczyk, Gable, Lyon, Junge, Wyder,
  Huerta-Cepas, Simonovic, Doncheva, Morris, Bork, Jensen and
  Mering}]{10.1093/nar/gky1131}
\bibinfo{author}{Szklarczyk, D.}, \bibinfo{author}{Gable, A.L.},
  \bibinfo{author}{Lyon, D.}, \bibinfo{author}{Junge, A.},
  \bibinfo{author}{Wyder, S.}, \bibinfo{author}{Huerta-Cepas, J.},
  \bibinfo{author}{Simonovic, M.}, \bibinfo{author}{Doncheva, N.T.},
  \bibinfo{author}{Morris, J.H.}, \bibinfo{author}{Bork, P.},
  \bibinfo{author}{Jensen, L.J.}, \bibinfo{author}{Mering, C.},
  \bibinfo{year}{2019}.
\newblock \bibinfo{title}{{STRING v11}: protein-protein association networks
  with increased coverage, supporting functional discovery in genome-wide
  experimental datasets}.
\newblock \bibinfo{journal}{Nucleic Acids Research} \bibinfo{volume}{47},
  \bibinfo{pages}{D607--D613}.
\newblock \DOIprefix\doi{10.1093/nar/gky1131}.
%Type = Article
\bibitem[{Tang et~al.(2024)Tang, Li, Li, Cao, Liu and
  Zeng}]{10.1093/bioinformatics/btae603}
\bibinfo{author}{Tang, T.}, \bibinfo{author}{Li, T.}, \bibinfo{author}{Li, W.},
  \bibinfo{author}{Cao, X.}, \bibinfo{author}{Liu, Y.}, \bibinfo{author}{Zeng,
  X.}, \bibinfo{year}{2024}.
\newblock \bibinfo{title}{Anti-symmetric framework for balanced learning of
  protein-protein interactions}.
\newblock \bibinfo{journal}{Bioinformatics} \bibinfo{volume}{40},
  \bibinfo{pages}{btae603}.
\newblock \DOIprefix\doi{10.1093/bioinformatics/btae603}.
%Type = Inproceedings
\bibitem[{{Vaswani} et~al.(2017){Vaswani}, {Shazeer}, {Parmar}, {Uszkoreit},
  {Jones}, {Gomez}, {Kaiser} and {Polosukhin}}]{vaswani2017attention}
\bibinfo{author}{{Vaswani}, A.}, \bibinfo{author}{{Shazeer}, N.},
  \bibinfo{author}{{Parmar}, N.}, \bibinfo{author}{{Uszkoreit}, J.},
  \bibinfo{author}{{Jones}, L.}, \bibinfo{author}{{Gomez}, A.N.},
  \bibinfo{author}{{Kaiser}, L.}, \bibinfo{author}{{Polosukhin}, I.},
  \bibinfo{year}{2017}.
\newblock \bibinfo{title}{Attention is all you need}, in:
  \bibinfo{booktitle}{Proceedings of the 31st International Conference on
  Neural Information Processing Systems}, pp. \bibinfo{pages}{5998--6008}.
%Type = Inproceedings
\bibitem[{Veličković et~al.(2018)Veličković, Cucurull, Casanova, Romero,
  Liò and Bengio}]{veličković2018graph}
\bibinfo{author}{Veličković, P.}, \bibinfo{author}{Cucurull, G.},
  \bibinfo{author}{Casanova, A.}, \bibinfo{author}{Romero, A.},
  \bibinfo{author}{Liò, P.}, \bibinfo{author}{Bengio, Y.},
  \bibinfo{year}{2018}.
\newblock \bibinfo{title}{Graph attention networks}, in:
  \bibinfo{booktitle}{Proceedings of the Sixth International Conference on
  Learning Representations}, pp. \bibinfo{pages}{1--12}.
\newblock \URLprefix \url{https://openreview.net/forum?id=rJXMpikCZ}.
%Type = Article
\bibitem[{Wang et~al.(2022)Wang, Wu, Lu, Jiang, Huang and
  Cai}]{Wang2022Protein}
\bibinfo{author}{Wang, S.}, \bibinfo{author}{Wu, R.}, \bibinfo{author}{Lu, J.},
  \bibinfo{author}{Jiang, Y.}, \bibinfo{author}{Huang, T.},
  \bibinfo{author}{Cai, Y.D.}, \bibinfo{year}{2022}.
\newblock \bibinfo{title}{Protein-protein interaction networks as miners of
  biological discovery}.
\newblock \bibinfo{journal}{Proteomics} \bibinfo{volume}{22},
  \bibinfo{pages}{2100190}.
\newblock \DOIprefix\doi{10.1002/pmic.202100190}.
%Type = Article
\bibitem[{Wang and Qi(2023)}]{9873966}
\bibinfo{author}{Wang, X.}, \bibinfo{author}{Qi, G.J.}, \bibinfo{year}{2023}.
\newblock \bibinfo{title}{Contrastive learning with stronger augmentations}.
\newblock \bibinfo{journal}{IEEE Transactions on Pattern Analysis and Machine
  Intelligence} \bibinfo{volume}{45}, \bibinfo{pages}{5549--5560}.
\newblock \DOIprefix\doi{10.1109/TPAMI.2022.3203630}.
%Type = Article
\bibitem[{Wong et~al.(2016)Wong, You, Ming, Li, Chen and Huang}]{ijms17010021}
\bibinfo{author}{Wong, L.}, \bibinfo{author}{You, Z.H.}, \bibinfo{author}{Ming,
  Z.}, \bibinfo{author}{Li, J.}, \bibinfo{author}{Chen, X.},
  \bibinfo{author}{Huang, Y.A.}, \bibinfo{year}{2016}.
\newblock \bibinfo{title}{Detection of interactions between proteins through
  rotation forest and local phase quantization descriptors}.
\newblock \bibinfo{journal}{International Journal of Molecular Sciences}
  \bibinfo{volume}{17}.
\newblock \DOIprefix\doi{10.3390/ijms17010021}.
%Type = Inproceedings
\bibitem[{Wu et~al.(2024)Wu, Tian, Huang, Li, Lin, Chawla and
  Li}]{wu2024mapeppi}
\bibinfo{author}{Wu, L.}, \bibinfo{author}{Tian, Y.}, \bibinfo{author}{Huang,
  Y.}, \bibinfo{author}{Li, S.}, \bibinfo{author}{Lin, H.},
  \bibinfo{author}{Chawla, N.V.}, \bibinfo{author}{Li, S.Z.},
  \bibinfo{year}{2024}.
\newblock \bibinfo{title}{{MAPE-PPI}: Towards effective and efficient
  protein-protein interaction prediction via microenvironment-aware protein
  embedding}, in: \bibinfo{booktitle}{Proceedings of the Twelfth International
  Conference on Learning Representations}, pp. \bibinfo{pages}{1--16}.
%Type = Article
\bibitem[{Wu et~al.(2022)Wu, Sun, Zhang, Xie and Cui}]{10.1145/3535101}
\bibinfo{author}{Wu, S.}, \bibinfo{author}{Sun, F.}, \bibinfo{author}{Zhang,
  W.}, \bibinfo{author}{Xie, X.}, \bibinfo{author}{Cui, B.},
  \bibinfo{year}{2022}.
\newblock \bibinfo{title}{Graph neural networks in recommender systems: A
  survey}.
\newblock \bibinfo{journal}{ACM Computing Surveys} \bibinfo{volume}{55}.
\newblock \DOIprefix\doi{10.1145/3535101}.
%Type = Inproceedings
\bibitem[{Xu et~al.(2019)Xu, Hu, Leskovec and Jegelka}]{xu2018how}
\bibinfo{author}{Xu, K.}, \bibinfo{author}{Hu, W.}, \bibinfo{author}{Leskovec,
  J.}, \bibinfo{author}{Jegelka, S.}, \bibinfo{year}{2019}.
\newblock \bibinfo{title}{How powerful are graph neural networks?}, in:
  \bibinfo{booktitle}{Proceedings of the Seventh International Conference on
  Learning Representations}, pp. \bibinfo{pages}{1--17}.
\newblock \URLprefix \url{https://openreview.net/forum?id=ryGs6iA5Km}.
%Type = Article
\bibitem[{Yao et~al.(2019)Yao, Du, Diao and Zhu}]{10.7717/peerj.7126}
\bibinfo{author}{Yao, Y.}, \bibinfo{author}{Du, X.}, \bibinfo{author}{Diao,
  Y.}, \bibinfo{author}{Zhu, H.}, \bibinfo{year}{2019}.
\newblock \bibinfo{title}{An integration of deep learning with feature
  embedding for protein-protein interaction prediction}.
\newblock \bibinfo{journal}{PeerJ} \bibinfo{volume}{7}, \bibinfo{pages}{e7126}.
\newblock \URLprefix \url{https://doi.org/10.7717/peerj.7126},
  \DOIprefix\doi{10.7717/peerj.7126}.
%Type = Article
\bibitem[{You et~al.(2015)You, Chan and Hu}]{You2015Predicting}
\bibinfo{author}{You, Z.H.}, \bibinfo{author}{Chan, K.C.C.},
  \bibinfo{author}{Hu, P.}, \bibinfo{year}{2015}.
\newblock \bibinfo{title}{Predicting protein-protein interactions from primary
  protein sequences using a novel multi-scale local feature representation
  scheme and the random forest}.
\newblock \bibinfo{journal}{PLOS ONE} \bibinfo{volume}{10},
  \bibinfo{pages}{e0125811}.
\newblock \DOIprefix\doi{10.1371/journal.pone.0125811}.
%Type = Inproceedings
\bibitem[{Yuan et~al.(2023)Yuan, Zhang, Zhou, Wang, Qiu, Shao, Zhang, Long,
  Kuang, Yao, Han, Ding, Lin, Wu and Wang}]{NEURIPS2023_9ed1c94a}
\bibinfo{author}{Yuan, J.}, \bibinfo{author}{Zhang, X.}, \bibinfo{author}{Zhou,
  H.}, \bibinfo{author}{Wang, J.}, \bibinfo{author}{Qiu, Z.},
  \bibinfo{author}{Shao, Z.}, \bibinfo{author}{Zhang, S.},
  \bibinfo{author}{Long, S.}, \bibinfo{author}{Kuang, K.},
  \bibinfo{author}{Yao, K.}, \bibinfo{author}{Han, J.}, \bibinfo{author}{Ding,
  E.}, \bibinfo{author}{Lin, L.}, \bibinfo{author}{Wu, F.},
  \bibinfo{author}{Wang, J.}, \bibinfo{year}{2023}.
\newblock \bibinfo{title}{{HAP}: Structure-aware masked image modeling for
  human-centric perception}, in: \bibinfo{editor}{Oh, A.},
  \bibinfo{editor}{Naumann, T.}, \bibinfo{editor}{Globerson, A.},
  \bibinfo{editor}{Saenko, K.}, \bibinfo{editor}{Hardt, M.},
  \bibinfo{editor}{Levine, S.} (Eds.), \bibinfo{booktitle}{Proceedings of the
  Thirty-Seventh Conference on Neural Information Processing Systems},
  \bibinfo{publisher}{Curran Associates, Inc.}. pp.
  \bibinfo{pages}{50597--50616}.
%Type = Article
\bibitem[{Zeng et~al.(2024)Zeng, Meng, Wen, Li and Li}]{Zeng2024GNNGL-PPI}
\bibinfo{author}{Zeng, X.}, \bibinfo{author}{Meng, F.F.}, \bibinfo{author}{Wen,
  M.L.}, \bibinfo{author}{Li, S.J.}, \bibinfo{author}{Li, Y.},
  \bibinfo{year}{2024}.
\newblock \bibinfo{title}{{GNNGL-PPI}: multi-category prediction of
  protein-protein interactions using graph neural networks based on global
  graphs and local subgraphs}.
\newblock \bibinfo{journal}{BMC Genomics} \bibinfo{volume}{25},
  \bibinfo{pages}{406}.
\newblock \DOIprefix\doi{10.1186/s12864-024-10299-x}.
%Type = Inproceedings
\bibitem[{Zhang et~al.(2019)Zhang, Song, Huang, Swami and
  Chawla}]{10.1145/3292500.3330961}
\bibinfo{author}{Zhang, C.}, \bibinfo{author}{Song, D.},
  \bibinfo{author}{Huang, C.}, \bibinfo{author}{Swami, A.},
  \bibinfo{author}{Chawla, N.V.}, \bibinfo{year}{2019}.
\newblock \bibinfo{title}{Heterogeneous graph neural network}, in:
  \bibinfo{booktitle}{Proceedings of the 25th ACM SIGKDD International
  Conference on Knowledge Discovery \& Data Mining},
  \bibinfo{publisher}{Association for Computing Machinery},
  \bibinfo{address}{New York, NY, USA}. pp. \bibinfo{pages}{793--803}.
\newblock \DOIprefix\doi{10.1145/3292500.3330961}.
%Type = Inproceedings
\bibitem[{Zhang et~al.(2023)Zhang, Xu, Jamasb, Chenthamarakshan, Lozano, Das
  and Tang}]{zhang2023protein}
\bibinfo{author}{Zhang, Z.}, \bibinfo{author}{Xu, M.}, \bibinfo{author}{Jamasb,
  A.R.}, \bibinfo{author}{Chenthamarakshan, V.}, \bibinfo{author}{Lozano, A.},
  \bibinfo{author}{Das, P.}, \bibinfo{author}{Tang, J.}, \bibinfo{year}{2023}.
\newblock \bibinfo{title}{Protein representation learning by geometric
  structure pretraining}, in: \bibinfo{booktitle}{Proceedings of the Eleventh
  International Conference on Learning Representations}, pp.
  \bibinfo{pages}{1--27}.
\newblock \URLprefix \url{https://openreview.net/forum?id=to3qCB3tOh9}.
%Type = Inproceedings
\bibitem[{Zhao et~al.(2023)Zhao, Qian, Yang, Zeng, Guan, Tam and
  Li}]{zhao2023semignn}
\bibinfo{author}{Zhao, Z.}, \bibinfo{author}{Qian, P.}, \bibinfo{author}{Yang,
  X.}, \bibinfo{author}{Zeng, Z.}, \bibinfo{author}{Guan, C.},
  \bibinfo{author}{Tam, W.L.}, \bibinfo{author}{Li, X.}, \bibinfo{year}{2023}.
\newblock \bibinfo{title}{{SemiGNN-PPI}: self-ensembling multi-graph neural
  network for efficient and generalizable protein-protein interaction
  prediction}, in: \bibinfo{booktitle}{Proceedings of the Thirty-Second
  International Joint Conference on Artificial Intelligence}, pp.
  \bibinfo{pages}{4984--4992}.
\newblock \DOIprefix\doi{10.24963/ijcai.2023/554}.
%Type = Article
\bibitem[{Zhou et~al.(2017)Zhou, Yu, Ding, Guo and Gong}]{Zhou2017Multi-scale}
\bibinfo{author}{Zhou, C.}, \bibinfo{author}{Yu, H.}, \bibinfo{author}{Ding,
  Y.}, \bibinfo{author}{Guo, F.}, \bibinfo{author}{Gong, X.J.},
  \bibinfo{year}{2017}.
\newblock \bibinfo{title}{Multi-scale encoding of amino acid sequences for
  predicting protein interactions using gradient boosting decision tree}.
\newblock \bibinfo{journal}{PLOS ONE} \bibinfo{volume}{12},
  \bibinfo{pages}{1--18}.
\newblock \DOIprefix\doi{10.1371/journal.pone.0181426}.
%Type = Article
\bibitem[{Zhu et~al.(2024)Zhu, Darefsky and Duan}]{10731549}
\bibinfo{author}{Zhu, G.}, \bibinfo{author}{Darefsky, J.},
  \bibinfo{author}{Duan, Z.}, \bibinfo{year}{2024}.
\newblock \bibinfo{title}{Cacophony: An improved contrastive audio-text model}.
\newblock \bibinfo{journal}{IEEE/ACM Transactions on Audio, Speech, and
  Language Processing} \bibinfo{volume}{32}, \bibinfo{pages}{4867--4879}.
\newblock \DOIprefix\doi{10.1109/TASLP.2024.3485170}.
%Type = Article
\bibitem[{Zhu et~al.(2001)Zhu, Bilgin, Bangham, Hall, Casamayor, Bertone, Lan,
  Jansen, Bidlingmaier, Houfek, Mitchell, Miller, Dean, Gerstein and
  Snyder}]{Zhu2001Global}
\bibinfo{author}{Zhu, H.}, \bibinfo{author}{Bilgin, M.},
  \bibinfo{author}{Bangham, R.}, \bibinfo{author}{Hall, D.},
  \bibinfo{author}{Casamayor, A.}, \bibinfo{author}{Bertone, P.},
  \bibinfo{author}{Lan, N.}, \bibinfo{author}{Jansen, R.},
  \bibinfo{author}{Bidlingmaier, S.}, \bibinfo{author}{Houfek, T.},
  \bibinfo{author}{Mitchell, T.}, \bibinfo{author}{Miller, P.},
  \bibinfo{author}{Dean, R.A.}, \bibinfo{author}{Gerstein, M.},
  \bibinfo{author}{Snyder, M.}, \bibinfo{year}{2001}.
\newblock \bibinfo{title}{Global analysis of protein activities using proteome
  chips}.
\newblock \bibinfo{journal}{Science} \bibinfo{volume}{293},
  \bibinfo{pages}{2101--2105}.
\newblock \DOIprefix\doi{10.1126/science.1062191}.

\end{thebibliography}

\end{document}